%% file: camera_ready.tex
\documentclass[final]{cvpr}
\usepackage{times}
\usepackage{epsfig}
\usepackage{graphicx}
\usepackage{amsmath}
\usepackage{amssymb}
\usepackage{subfiles}

\usepackage{color,soul}
\soulregister\cite7
\soulregister\ref7
\soulregister\pageref7

\usepackage[pagebackref=true,breaklinks=true,colorlinks,bookmarks=false]{hyperref}

\def\cvprPaperID{2831} 
\def\confYear{CVPR 2021}

\begin{document}

\title{Monocular 3D Multi-Person Pose Estimation by Integrating \\ Top-Down and Bottom-Up Networks}


\author{
Yu Cheng$^{1}$, Bo Wang$^{2}$, Bo Yang$^{2}$, Robby T. Tan$^{1,3}$ \\
$^{1}$National University of Singapore\\ 
$^{2}$Tencent Game AI Research Center\\
$^{3}$Yale-NUS College\\
{\tt\small e0321276@u.nus.edu, \{bohawkwang,brandonyang\}@tencent.com, robby.tan@nus.edu.sg}
}

\maketitle

\begin{abstract}

In monocular video 3D multi-person pose estimation, inter-person occlusion and close interactions can cause human detection to be erroneous and human-joints grouping to be unreliable.
Existing top-down methods rely on human detection and thus suffer from these problems.
Existing bottom-up methods do not use human detection, but they process all persons at once at the same scale, causing them to be sensitive to multiple-persons scale variations. 
To address these challenges, we propose the integration of top-down and bottom-up approaches to exploit their strengths. 
Our top-down network estimates human joints from all persons instead of one in an image patch, making it robust to possible erroneous bounding boxes. 
Our bottom-up network incorporates human-detection based normalized heatmaps, allowing the network to be more robust in handling scale variations.
Finally, the estimated 3D poses from the top-down and bottom-up networks are fed into our integration network for final 3D poses.
Besides the integration of top-down and bottom-up networks, unlike existing pose discriminators that are designed solely for a single person, and consequently cannot assess natural inter-person interactions, we propose a two-person pose discriminator that enforces natural two-person interactions.
Lastly, we also apply a semi-supervised method to overcome the 3D ground-truth data scarcity.
Quantitative and qualitative evaluations show the effectiveness of the proposed method. 
%
Our code is available publicly. \footnote{\url{https://github.com/3dpose/3D-Multi-Person-Pose}}
\end{abstract}

\vspace{-0.5em}
\section{Introduction}
\begin{figure}[t]
	\centering
	\includegraphics[width=\linewidth]{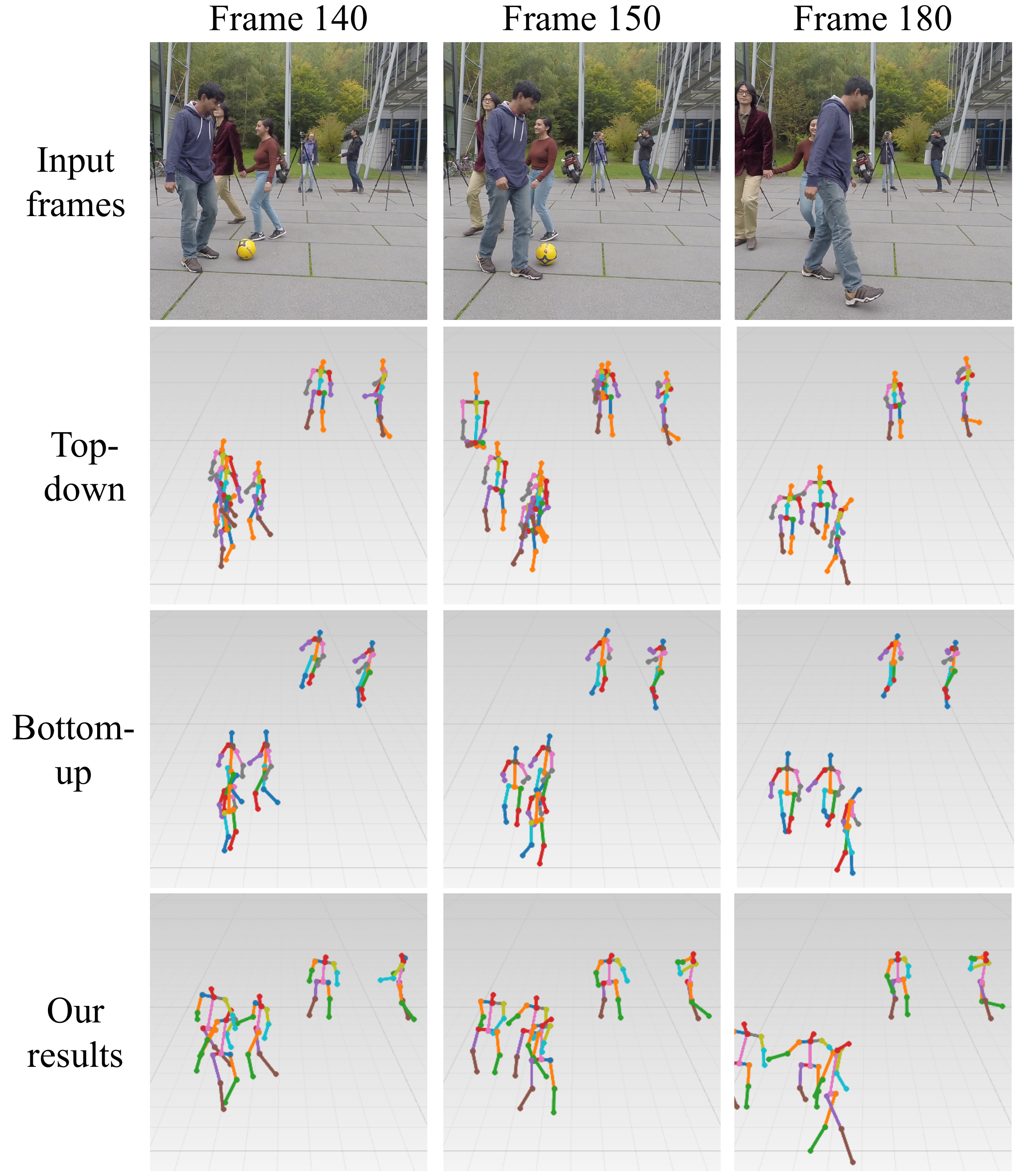}
	\vspace{-0.5em}
	\caption{Incorrect 3D multi-person pose estimation from existing top-down (2nd row) and bottom-up (3rd row) methods. The top-down method is RootNet~\cite{Moon_2019_ICCV_3DMPPE}, the bottom-up method is SMAP~\cite{zhen2020smap}. The input images are from MuPoTS-3D dataset~\cite{mehta2018single}. The top-down method suffers from inter-person occlusion and the bottom-up method is sensitive to scale variations (i.e., the 3D poses of the two persons in the back are inaccurately estimated). Our method substantially outperforms the state-of-the-art.}
	\label{fig:intriguing_example}
	\vspace{-1.0em}
\end{figure}

Estimating 3D multi-person poses from a monocular video has drawn increasing attention due to its importance for real-world applications (e.g., \cite{Moon_2019_ICCV_3DMPPE,mehta2020xnect,bertoni2019monoloco,cheng2020sptaiotemporal}). Unfortunately, it is generally still challenging and an open problem, particularly when multiple persons are present in the scene. Multiple persons can generate  inter-person occlusion, which causes human detection to be erroneous. Moreover, multiple persons in a scene are likely in close contact with each other and interact, which makes human-joints grouping unreliable.

Although existing 3D human pose estimation methods (e.g.,  \cite{mehta2017vnect,Yang20183DHP,pavlakos2018ordinal,hossain2018exploiting,pavllo20193d,cheng2019occlusion,cheng2020sptaiotemporal}) show promising results on single-person datasets like Human3.6M \cite{human36ionescu} and HumanEva \cite{sigal2010humaneva}, these methods do not perform well in 3D multi-person scenarios.
Generally, we can divide existing methods into two approaches: top-down and bottom-up.
Existing top-down 3D pose estimation methods rely considerably on human detection to localize each person, prior to estimating the joints within the detected bounding boxes, e.g., \cite{pavllo20193d,cheng2019occlusion,Moon_2019_ICCV_3DMPPE}. These methods show promising performance for single-person 3D-pose estimation \cite{pavllo20193d,cheng2019occlusion}, yet since they treat each person individually, they have no awareness of non-target persons and the possible interactions. When multiple persons occlude each other, human detection also become unreliable. Moreover, when target persons are closely interacting with each other, the pose estimator may be misled by the nearby persons, e.g., predicted joints may come from the nearby non-target persons.

Recent bottom-up methods (e.g., \cite{zhen2020smap,lin2020hdnet,li2020hmor}) do not use any human detection and thus can produce results with higher accuracy when multiple persons interact with each other.
These methods consider multiple persons simultaneously and, in many cases, better distinguish the joints of different persons. 
Unfortunately, without using detection, bottom-up methods suffer from the scale variations, and the pose estimation accuracy is compromised, rendering inferior performance compared with  top-down approaches \cite{cheng2020higherhrnet}.
As shown in Figure \ref{fig:intriguing_example}, neither top-down nor bottom-up approach alone can handle all the challenges at once, particularly the challenges of: inter-person occlusion, close interactions, and human-scale variations.
Therefore, in this paper, our goal is to integrate the top-down and bottom-up approaches to achieve more accurate and robust 3D multi-person pose estimation from a monocular video.

To achieve this goal, we introduce a top-down network to estimate human joints inside each detected bounding box. 
%
Unlike existing top-down methods that only estimate one human pose given a bounding box, our top-down network predicts 3D poses for all persons inside the bounding box. 
%
%
%
The joint heatmaps from our top-down network is feed to our bottom-up network, so that our bottom network can be more robust in handling the scale variations.
Finally, we feed the estimated 3D poses from both top-down and bottom-up networks into our integration network to obtain the final estimated 3D poses given an image sequence.

Moreover, unlike existing methods' pose discriminators, which are designed solely for single person, and consequently cannot enforce natural inter-person interactions, we propose a two-person pose discriminator that enforces two-person natural interactions. Lastly, semi-supervised learning is used to mitigate the data scarcity problem where 3D ground-truth data is limited. 

In summary, our contributions are listed as follows. 
\begin{itemize}
    \item We introduce a novel two-branch framework, where the top-down branch detects multiple persons and the bottom-up branch incorporates the normalized image patches in its process. Our framework gains benefits from the two branches, and at the same time, overcomes their shortcomings. 
	\item We employ multi-person pose estimation for our top-down network, which can effectively handle the inter-person occlusion and interactions caused by detection errors. 
	\item We incorporate human detection information into our bottom-up branch so that  it can better handle the scale variation, which addresses the problem in existing bottom-up methods.
	\item Unlike the existing discriminators that focus on single person pose, we introduce a novel discriminator that enforces the validity of human poses of close pairwise interactions in the camera-centric coordinates.
\end{itemize}

\section{Related Works}

\noindent \textbf{Top-Down Monocular 3D Human Pose Estimation} 
Existing top-down 3D human pose estimation methods commonly use human detection as an essential part of their methods to estimate person-centric 3D human poses~\cite{martinez2017simple,nie2017monocular,mehta2017vnect,pavllo20193d,cheng2019occlusion,doersch2019sim2real,cheng2020sptaiotemporal}. 
They demonstrate promising performance on single-person evaluation datasets~\cite{human36ionescu,sigal2010humaneva}, unfortunately the performance decreases in multi-person scenarios, due to inter-person occlusion or close interactions \cite{mehta2017vnect,cheng2019occlusion}. Moreover, the produced person-centric 3D poses cannot be used for multi-person scenarios, where camera-centric 3D-pose estimation is needed.
Top-down methods process each person independently, leading to inadequate awareness of the existence of other persons nearby. 
As a result, they perform poorly on multi-person videos where inter-person occlusion and close interactions  are commonly present. 
Rogez et al.~\cite{rogez2017lcr,rogez2019lcr} develop a pose proposal network to generate bounding boxes and then perform pose estimation individually for each person. Recently, unlike previous methods that perform person-centric pose estimation, Moon et al.~\cite{Moon_2019_ICCV_3DMPPE} propose a top-down 3D multi-person pose-estimation method that can estimate the poses for all persons in an image in the camera-centric coordinates. However, the method still relies on detection and process each person independently; hence it is likely to suffer from inter-person occlusion and close interactions.

\begin{figure*}[h]
    \centering
    \includegraphics[width=\textwidth]{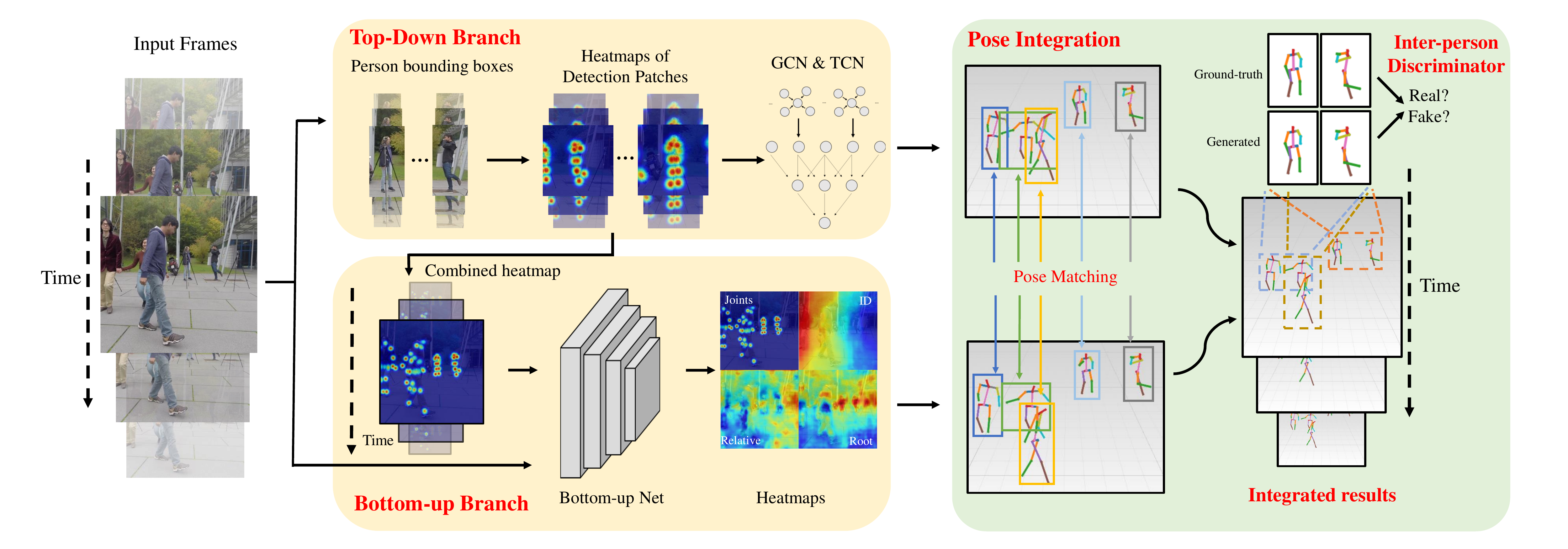}
    \caption{The overview of our framework. Our proposed method comprises three components: 1) A top-down branch to estimate fine-grained instance-wise 3D pose. 2) A bottom-up branch to generate global-aware camera-centric 3D pose. 3) An integration network to generate final estimation based on paired poses from top-down and bottom-up to take benefits from both branches. Note that the semi-supervised learning part is a training strategy so it is not included in this figure.
    }
    \label{fig:pipeline}
\end{figure*}

\vspace{0.1cm}
\noindent \textbf{Bottom-Up Monocular 3D Human Pose Estimation} 
A few bottom-up methods have been proposed~\cite{fabbri2020compressed,zhen2020smap,mehta2020xnect,li2020hmor,lin2020hdnet}. Fabbri et al.~\cite{fabbri2020compressed} introduce an encoder-decoder framework to compress a heatmap first, and then decompress it back to the original representations in the test time for fast HD image processing. 
Mehta et al.~\cite{mehta2020xnect} propose to identify individual joints, compose full-body joints, and enforce temporal and kinematic constraints in three stages for real-time 3D motion capture. 
Li et al.~\cite{li2020hmor} develop an integrated method with lower computation complexity for human detection, person-centric pose estimation, and human depth estimation from an input image. 
Lin et al.~\cite{lin2020hdnet}  formulate the human depth regression as a bin index estimation problem for multi-person localization in the camera coordinate system.
Zhen et al.~\cite{zhen2020smap} estimate the 2.5D representation of body parts first and then reconstruct camera-centric multi-person 3D poses. 
These methods benefit from the nature of the bottom-up approach, which can process multiple persons simultaneously without relying on human detection.
However, since all persons are processed at the same scale, these methods are inevitably sensitive to human scale variations, which limits their applicability on wild videos. 

\vspace{0.1cm}
\noindent \textbf{Top-Down and Bottom-Up Combination} 
Earlier non-deep learning methods exploring the combination of top-down and bottom-up approaches for human pose estimation are in the forms of data-driven belief propagation, different classifiers for joint location and skeleton, or probabilistic Gaussian mixture modelling~\cite{hua2005learning,wang2010combined,kuo2011integration}. 
Recent deep learning based methods that attempt to make use of both top-down and bottom-up information are mainly on estimating 2D poses~\cite{hu2016bottom,tang2018deeply,cai2019exploiting,li2019multi}. 
Hu and Ramanan~\cite{hu2016bottom} propose a hierarchical rectified Gaussian model to incorporate top-down feedback with bottom-up CNNs. 
Tang et al.~\cite{tang2018deeply} develop a framework with bottom-up inference followed by top-down refinement based on a compositional model of the human body.
Cai et al.~\cite{cai2019exploiting} introduce a spatial-temporal graph convolutional network (GCN) that uses both bottom-up and top-down features. 
These methods explore to benefit from top-down and bottom-up information. However, 
they are not suitable for 3D multi-person pose estimation because the fundamental weaknesses in both top-down and bottom-up methods are not addressed completely, which include inter-person occlusion caused detection and joints grouping errors, and the scale variation issue. 
Li et al.~\cite{li2019multi} adopt LSTM and combine bottom-up heatmaps with human detection for 2D multi-person pose estimation. They address occlusion and detection shift problems. Unfortunately, they use a bottom-up network and only add the detection bounding box as the top-down information to group the joints. Hence, their method is essentially still bottom-up and thus still vulnerable to human scale variations.

\section{Proposed Method}

Fig. \ref{fig:pipeline} shows our pipeline, which consists of three major parts to accomplish the multi-person camera-centric 3D human pose estimation: a top-down network for fine-grained instance-wise pose estimation, a bottom-up network for global-aware pose estimation, and an integration network to integrate the estimations of top-down and bottom-up branches with inter-person pose discriminator. Moreover, a semi-supervised training process is proposed to enhance the 3D pose estimation based on reprojection consistency. 

\subsection{Top-Down Network}
\label{sec:top-down}

\begin{figure}
    \centering
    \includegraphics[width=0.9\linewidth]{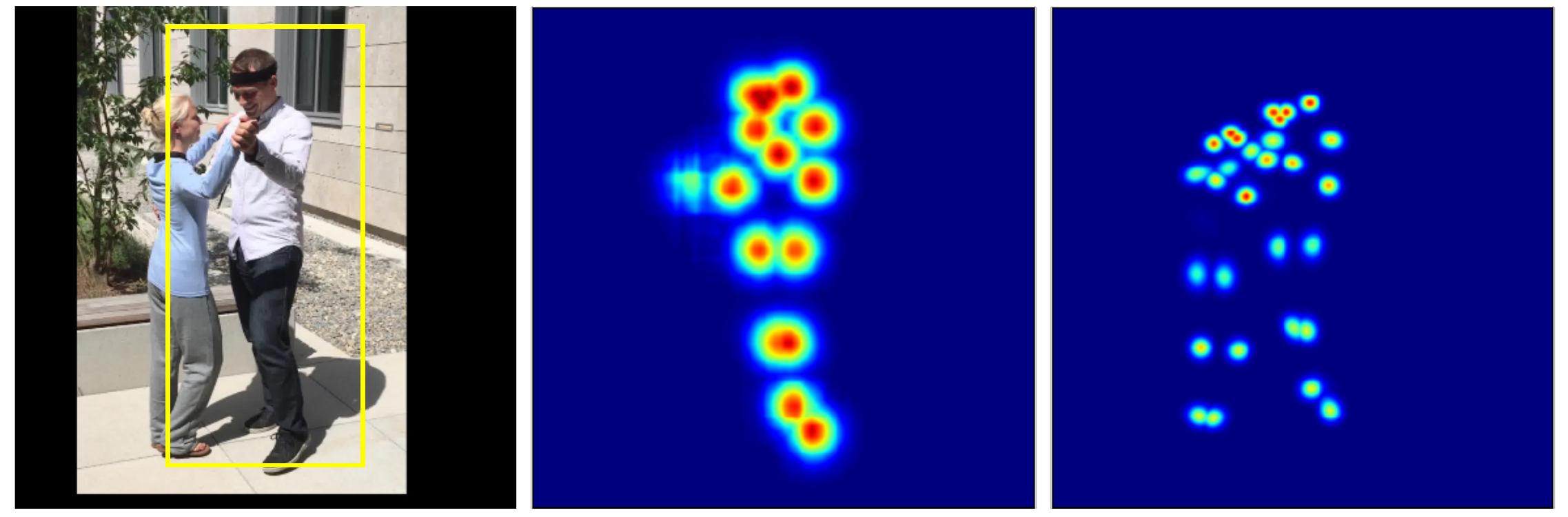}
    \caption{Examples of estimated heatmaps of human joints. 
    The left image shows the input frame overlaid with inaccurate detection bounding box (i.e., only one person detected).
    The middle image shows the estimated heatmap of existing top-down methods. The right image shows the heatmap of our top-down branch.
    }
    \label{fig:topdownfail}
\end{figure}

Given a human detection bounding box, existing top-down methods  estimate full-body joints of one person. 
Consequently, if there are multiple persons inside the box or partially out-of-bounding box body parts, the full-body joint estimation are likely to be erroneous.
Figure \ref{fig:topdownfail} shows such failure examples of existing methods. 
In contrast, our method produces the heatmaps for all joints inside the bounding box (i.e., enlarged to accommodate inaccurate detection), and estimate the ID for each joint to group them into corresponding persons, similar to~\cite{newell2017associative}. 

Given an input video, for every frame we apply a human detector \cite{he2017mask}, and crop the image patches based on the detected bounding boxes.
A 2D pose detector \cite{cheng2020higherhrnet} is applied to each patch to generate heatmaps for all human joints, such as shoulder, pelvis, ankle, and etc. 
Specifically, our top-down loss of 2D pose heatmap is an L2 loss between the predicted and ground-truth heatmaps, formulated as:
\begin{equation}
    L^{TD}_{hmap} = |H - \Tilde{H}|^2_2, 
\label{eq:hmap}
\end{equation}
where $H$ and $\Tilde{H}$ are the predicted and ground-truth heatmaps, respectively. 

Having obtained the 2D pose heatmaps, a directed GCN network is used to refine the potentially incomplete poses caused by occlusions or partially out-of-bounding box body parts, and two TCNs are used to estimate both person-centric 3D pose and camera-centric root depth based on a given sequence of 2D poses similar to~\cite{cheng2021graph}. As the TCN requires the input sequence of the same instance, a pose tracker \cite{umer2020self} is used to track each instance in the input video. We also apply data augmentation in training our TCN so that it can handle occlusions \cite{cheng2019occlusion}. 

\subsection{Bottom-Up Network}
\label{sec:bottom-up}

Top-down methods perform estimation inside the bounding boxes, and thus are lack of global awareness of other persons, leading to difficulties to estimate poses in the camera-centric coordinates. 
To address this problem, we further propose a bottom-up network that processes multiple persons simultaneously.
Since the bottom-up pose estimation suffers from human scale variations, we concatenate the heatmaps from our top-down network with the original input frame as the input of our bottom-up network.
With the guidance of the top-down heatmaps, which are the results of the object detector and pose estimation based on  the normalized boxes, the estimation of the bottom-up network will be more robust to scale variations.
Our bottom-up network outputs four heatmaps : a 2D pose heatmap, ID-tag map, relative depth map, and root depth map. 
The 2D pose heatmap and ID-tag map are defined in the same way as in the previous section (\ref{sec:top-down}). 
The relative depth map refers to the depth map of each joint with respect to its root (pelvis) joint. 
The root depth map represents the depth map of the root joint. 

In particular, the loss functions $L^{BU}_{hmap}$ and $L^{BU}_{id}$ for the heatmap and ID-tag map are similar to \cite{newell2017associative}. 
In addition, we apply the depth loss to the estimations of both the relative depth map $h^{rel}$ and the root depth $h^{root}$. Please see supplementary material for example of the four estimated heatmaps from the bottom-up network. For $N$ persons and $K$ joints, the loss can be formulated as:
\begin{equation}
    L_{depth} = \frac{1}{NK}\sum_n \sum_k |h_{k}(x_{nk}, y_{nk}) - d_{nk}|^2,
\end{equation}
where $h$ is the depth map and $d$ is the ground-truth depth value. 
Note that, for pelvis (i.e., the root joint), the depth is a camera-centric depth.
For other joints, the depth is relative with respect to the corresponding root joint. 

We group the heatmaps into instances (i.e., persons), and retrieve the joint locations using the same procedure as in the top-down network. 
Moreover, the values of the camera-centric depth of the root joint $z^{root}$ and the relative depth for the other joints $z^{rel}_k$ are obtained by retrieving from the corresponding depth maps where the joints (i.e., root or others) are located.
Specifically:
\begin{eqnarray}
    z^{root}_{i} &=& h^{root}(x^{root}_i, y^{root}_i) \\
    z^{rel}_{i,k} &=& h^{rel}_k (x_{i,k}, y_{i,k})
\end{eqnarray}
where $i,k$ refer to the $i_{th}$ instance and $k_{th}$ joint, respectively.

\subsection{Integration  with Interaction-Aware Discriminator}

Having obtained the results from the top-down and bottom-up networks, 
we first need to find the corresponding poses between the results from the two networks, i.e., the top-down pose $P^{TD}_i$ and bottom-up pose $P^{BU}_j$ belong to the same person. Note that $P$ stands for camera-centric 3D pose throughout this paper. 

Given two pose sets from bottom-up branch $P^{BU}$ and top-down branch $P^{TD}$, we match the poses from both sets, in order to form pose pairs. 
The similarity of two poses is defined as:
\begin{eqnarray}
    {\rm Sim}_{i,j} = \sum_{k=0}^K \min(c^{BU}_{i,k}, c^{TD}_{j,k}) {\rm OKS} (P^{BU}_{i,k}, P^{TD}_{j,k}),
\end{eqnarray}
where:
\begin{eqnarray}
    {\rm OKS}(x,y) = \exp(- \frac{d(x,y)^2}{2s^2\sigma^2}),
\end{eqnarray}
${\rm OKS}$ stands for object keypoint similarity \cite{xiao2018simple}, which measures the joint similarity of a given joint pair. 
$d(x,y)$ is the Euclidean distance between two joints. 
$s$ and $\sigma$ are two controlling parameters. 
${\rm Sim}_{i,j}$ measures the similarity between the $i_{th}$ 3D pose $P^{BU}_{i}$ from the bottom-up network and the $j_{th}$ 3D pose $P^{TD}_{j}$ from the top-down network  over $K$ joints.
Note that both poses from top-down $P^{TD}$ and bottom-up $P^{BU}$ are camera-centric; thus, the similarity is measured based on the camera coordinate system. 
The $c^{BU}_{i,k}$ and $c^{TD}_{j,k}$ are the confidence values of joint $k$ for 3D poses $P^{BU}_{i}$ and $P^{TD}_{j}$, respectively. 
Having computed the similarity matrix between the two sets of poses $P^{TD}$ and $P^{BU}$ according to the ${\rm Sim}_{i,j}$ definition, the Hungarian algorithm \cite{kuhn1955hungarian} is used to obtain the matching results.

Once the matched pairs are obtained, we feed each pair of the 3D poses and the confidence score of each joint to our integration network.
Our integration network consists of 3 fully connected layers, which outputs the final estimation. 

\vspace{0.3cm}
\noindent \textbf{Integration Network Training}
To train the integration network, we take some samples from the ground-truth 3D poses.
We apply data augmentation: 
1) random masking the joints with a binary mask $M^{kpt}$ to simulate occlusions; 
2) random shifting the joints to simulate the inaccurate pose detection; 
and  3) random zeroing one from a pose pair to simulate unpaired poses. 
The loss of the integration network is an L2 loss between the predicted 3D pose and its ground-truth:
\begin{equation}
    L_{int} = \frac{1}{K} \sum_k |P_k - \Tilde{P}_k|^2, 
\end{equation}
where $K$ is the number of the estimated joints. $P$ and $\Tilde{P}$ are the estimated and ground-truth 3D poses, respectively.

\vspace{0.3cm}
\noindent \textbf{Inter-Person Discriminator}
For training the integration network, we propose a novel inter-person discriminator.
Unlike most existing discriminators  for human pose estimation (e.g. \cite{wandt2019repnet,cheng2020sptaiotemporal}), where they can only discriminate the plausible 3D poses of one person, we propose an interaction-aware discriminator to enforce the interaction of a pose pair is natural and reasonable, which not only includes the existing single-person discriminator, but also generalize to interacting persons. 
Specifically, our discriminator contains two sub-networks: $D_1$, which is dedicated for  one person-centric 3D poses; and, $D_2$, which is dedicated for a pair of camera-centric 3D poses from two persons. 
We apply the following loss to train the network, which is formulated as:
\begin{equation}
L_{dis} = log(\Tilde{C}) + log(1-C)
\end{equation}
where:
\begin{equation}
\begin{split}
    &C = 0.25 (D_1(P^a) + D_1(P^b)) +  0.5 D_2(P^a, P^b) \\
    &\Tilde{C} = 0.25 (D_1(\Tilde{P}^a) + D_1(\Tilde{P^b})) + 0.5 D_2(\Tilde{P}^a, \Tilde{P^b})
\end{split}
\end{equation}
where $P^a, P^b$ are the estimated poses of person $a$ and person $b$, respectively. $\Tilde{P}$ are the estimated and ground-truth 3D poses, respectively. 

\subsection{Semi-Supervised Training}

Semi-supervised learning is an effective technique to improve the network performance, particularly when the data with ground-truths are limited.
A few works also explore to make use of the unlabeled data~\cite{chen2019unsupervised,umer2020self,xu20203dssl}. 
In our method, we apply  a noisy student training strategy~\cite{xie2020self}.
We first train a teacher network with the 3D ground-truth dataset only, and then use the teacher network to generate their pseudo-labels of unlabelled data, which are used to  train a student network. 

The pseudo-labels cannot be directly used because some of them are likely incorrect. 
Unlike in the noisy student training strategy~\cite{xie2020self}, where data with ground-truth labels and pseudo-labels are mixed to train the student network by adding various types of noise (i. e., augmentations, dropout, etc), 
we propose  two-consistency loss terms to assess the quality of the pseudo-labels, including the reprojection error and multi-perspective error~\cite{chen2019unsupervised,pavllo20193d}.


The reprojection error measures the deviation between the projection of generated 3D poses and the detected 2D poses. Since there are more abundant data variations in 2D pose dataset compared to 3D pose dataset (e.g., COCO is much larger compared to H36M), the 2D estimator is expected to be more reliable than its 3D counterpart. 
Therefore, minimizing a reprojection error is helpful to improve the accuracy of 3D pose estimation.

The multi-perspective error, $E_{mp}$, measures the consistency of the predicted 3D poses from different viewing angles. This error indicates the reliability of the predicted 3D poses.  Based on the two terms,  our semi-supervised loss, $L_{\rm SSL}$, is formulated as,
\begin{equation}
    L_{\rm SSL} = w(E_{rep} + E_{mp}) + L_{dis},
\end{equation}
where $w$ is a weighting factor to balance the contribution of the reprojection and multi-perspective errors. 
In the training stage, $w$ first focuses on easy samples and gradually includes the hard samples. The weight, $w$, is  formulated as:
\begin{equation}
    w = {\rm softmax}(\frac{E_{rep}}{r}) + {\rm softmax}(\frac{E_{mp}}{r}),
\end{equation}
where $r$ is the number of training epochs. More details regarding to the reprojection and multi-perspective errors and the self-training process are discussed in the supplementary material.

\section{Experiment}

\paragraph{Datasets}
\label{sec:datasets}

We use MuPoTS-3D~\cite{mehta2018single} and JTA~\cite{fabbri2018learning} datasets to evaluate the camera-centric 3D multi-person pose estimation performance by following the existing methods~\cite{Moon_2019_ICCV_3DMPPE,fabbri2020compressed} and their training protocols (i.e., train, test split). In addition, we use 3DPW~\cite{3DPW} to evaluate person-centric 3D multi-person pose estimation performance following~\cite{humanMotionKanazawa19,sun2019human}. We also perform evaluation on the widely used Human3.6M dataset~\cite{human36ionescu} for person-centric 3D human pose estimation following~\cite{pavllo20193d,wandt2019repnet}. Details of the datasets information are in the supplementary material.

\vspace{-1.0em}
\paragraph{Implementation Details}

We use HRNet-w32 \cite{sun2019hrnet} as the backbone network for both multi-person pose estimator in the top-down and bottom-up networks. The top-down network is trained for $100$ epochs on the COCO dataset \cite{lin2014microsoft} with the Adam optimizer and learning rate $0.001$. The bottom-up network is trained for $50$ epochs with the Adam optimizer and learning rate $0.001$ on a combined dataset of MuCO \cite{singleshotmultiperson2018} and COCO \cite{lin2014microsoft}. More details are in the supplementary material. 




\vspace{-1.0em}
\paragraph{Evaluation Metrics}

\begin{table}[t]
	\footnotesize
	\centering
	\begin{tabular}{c|c|c|c|c}
		\cline{1-5}
		\rule{0pt}{2.6ex}
		\textbf{Method} & $AP_{25}^{root}$ & $AUC_{rel}$ & PCK & PCK$_{abs}$\\
		\cline{1-5}
		\rule{0pt}{2.6ex}
		TD (w/o MP) & 43.7 & 41.0 & 81.6 & 42.8\\
		TD (w MP) & 45.2 & 48.9 & 87.5 & 45.7\\
		BU (w/o CH) & 44.2 & 34.5 & 76.6 & 40.2\\
		BU (w CH) & \underline{46.1} & 35.1 & 78.0 & 41.5\\
		TD + BU (w/o MP,CH) & 44.9 & 42.6 & 82.8 & 43.1\\
		TD + BU (hard) & \underline{46.1} & 48.9 & 87.5 & 46.2\\
		TD + BU (linear) & \underline{46.1} & \underline{49.2} & \underline{88.0} & \underline{46.7}\\
		TD + BU (w/o PM) & 46.0 & 48.6 & 85.5 & 45.3 \\
		TD + BU (IN) & \textbf{46.3} & \textbf{49.6} & \textbf{88.9} & \textbf{47.4}\\
		\cline{1-5}
	\end{tabular}
	\vspace{0.5em}
	\caption{Ablation study on MuPoTS-3D dataset. TD, BU, MP, CH, IN, and PM stand for top-down, bottom-up, multi-person pose estimator, combined heatmap, integration network, and pose matching,  respectively. Best in \textbf{bold}, second best \underline{underlined}.}
	\label{tab:Ablation}
\end{table}

Since the majority of 3D human pose estimation methods produce person-centric 3D poses, to be able to compare, we perform person-centric 3D human pose estimation. We use Mean Per Joint Position Error (MPJPE), Procrustes analysis MPJPE (PA-MPJPE), Percentage of Correct 3D Keypoints (PCK), and area under PCK curve from various thresholds ($AUC_{rel}$) following the literature~\cite{Moon_2019_ICCV_3DMPPE,pavllo20193d,cheng2020sptaiotemporal}. Since we focus on 3D multi-person camera-centric pose estimation, we also use the metrics designed for evaluating performance in the camera coordinate system, including average precision of 3D human root location ($AP_{25}^{root}$) and PCK$_{abs}$, which is PCK without root alignment to evaluate the absolute camera-centric coordinates from~\cite{Moon_2019_ICCV_3DMPPE}, and F1 value following~\cite{fabbri2020compressed}. 

\vspace{-1.0em}
\paragraph{Ablation Studies}


Ablation studies are performed to validate the effectiveness of each sub-module of our framework. 
We validate our top-down network by using an existing top-down pose estimator (i.e.,  detection of one full-body joints) as a baseline, abbreviated as TD (w/o MP) to compare to our top-down network denoted as TD (w MP). 
We also validate our bottom-up network by using existing bottom-up heatmap estimation (i.e., estimate all person at the same scale) as a baseline, named BU (w/o CH) to compare to our bottom-up network, called BU (w CH). 
To evaluate our integration network, we use three baselines. 
The first is a straightforward integration by combining existing TD and BU networks. 
The second is hard integration, abbreviated TD + BU (hard), where the top-down person-centric pose is always used,  plus the root depth from the bottom-up network. 
The third is linear integration, abbreviated TD + BU (linear), where the person-centric 3D pose from top-down is combined with its corresponding bottom-up one based on the confidence values of the estimated heatmap. 

As shown in Table~\ref{tab:Ablation}, we observe that our top-down network, bottom-up network, and integration network clearly outperform their corresponding baselines. 
Our top-down network tends to have better person-centric 3D pose estimations compared with our bottom-up network, because the top-down network benefits from not only multi-person pose estimator, also GCN and TCN that help to deal with inter-occluded poses.
On the contrary, our bottom-up network achieves better performance for the root joint estimation, because it estimates the root depth based on a full image; while the root depth of top-down network is estimated based on an individual skeleton. 
Finally, our integration network demonstrates superior performance compared to hard or linear combining the poses from the top-down and bottom-up networks, which validates its effectiveness. 

Other than validating our top-down and bottom-up networks, we also perform ablation analysis on our semi-supervised learning. We show the result of using reprojection loss, multi-perspective loss, reprojection loss with our discriminator, and reprojection $\&$ multi-perspective loss with discriminator in Table~\ref{tab:ssl}. We can see that the reprojection loss is more useful than the multi-perspective loss because it leverages the information from the 2D pose estimator, which is trained with 2D datasets with a large number of  poses and environment variations. More importantly, we observe that our proposed interaction-aware discriminator makes the largest performance improvement compared with the other modules, demonstrating the importance of enforcing the validity of the interaction between persons. 
\begin{table}[t]
	\footnotesize
	\centering
	\begin{tabular}{c|c|c|c|c}
		\cline{1-5}
		\rule{0pt}{2.6ex}
		\textbf{Method} & $AP_{25}^{root}$ & $AUC_{rel}$ & PCK & PCK$_{abs}$\\
		\cline{1-5}
		\rule{0pt}{2.6ex}
		Rep & \textbf{46.3} & 43.4 & 77.2 & 40.7\\
		MP & \textbf{46.3} & 32.2 & 72.8 & 29.5\\
		Rep+dis & \textbf{46.3} & \underline{49.9} & \underline{89.1} & \underline{46.8}\\
        Rep+MP+dis & \textbf{46.3} & \textbf{50.6} & \textbf{89.6} & \textbf{48.0}\\
		\cline{1-5}
	\end{tabular}
	\vspace{0.5em}
	\caption{Ablation study on MuPoTS-3D dataset. Rep, MP, and dis stand for reprojection, multi-perspective, and discriminator. Best in \textbf{bold}, second best \underline{underlined}.  
	}
	\label{tab:ssl}
\end{table}

\vspace{-1.0em}
\paragraph{Quantitative Evaluation}

To evaluate the performance for 3D multi-person camera-centric pose estimation in both indoor and outdoor scenarios, we perform evaluations on MuPoTS-3D as summarized in Table~\ref{tab:MuPoTS_3d}. The results show that our camera-centric multi-person 3D pose estimation outperforms the SOTA \cite{li2020hmor} on $PCK_{abs}$ by $2.3\%$. We also perform person-centric 3D pose estimation evaluation using $PCK$ where we outperform the SOTA method \cite{lin2020hdnet} by $2.1\%$. The evaluation on MuPoTS-3D shows that our method outperforms the state-of-the-art methods 
in both camera-centric and person-centric 3D multi-person pose estimation as our framework overcomes the weaknesses of both bottom-up and top-down branches and at the same time benefits from their strengths.

Following recent work \cite{fabbri2020compressed}, we also perform evaluations on JTA, which is a synthetic dataset acquired from computer game, to further validate the effectiveness of our method for camera-centric 3D multi-person pose estimation. As shown in Table~\ref{tab:jta}, our method is superior over the SOTA method \cite{fabbri2020compressed} (e.g., our result shows $12.6\%$ improvement on F1 value, $t=0.4m$) on this challenging dataset where both inter-person occlusion and large person scale variation present, which again illustrate that our proposed method can handle these challenges in 3D multi-person pose estimation.

Human3.6M is widely used for evaluating 3D single-person pose estimation. As our method is focused on dealing with inter-person occlusion and scale variation, we do not expect our method performs significantly better than the SOTA methods. Table~\ref{tab:h36m} summarizes the quantitative evaluation on Human3.6M where our method is comparable with the SOTA methods \cite{kolotouros2019learning,li2020hmor} on person-centric 3D human pose evaluation metrics (i.e., MPJPE and PA-MPJPE). 

3DPW is an outdoor multi-person 3D human shape reconstruction dataset. It is unfair to compare the errors between skeleton-based method with ground-truth defined on SMPL model~\cite{loper2015smpl} due to the different definitions of joints~\cite{tripathi2020posenet3d}. We run human detection on all frames and create an occlusion subset where the frames with the large overlay between persons are selected. The performance drop between the full testing test of 3DPW and the occlusion subset can effectively tell if a method can handle inter-person occlusion, which is shown in Table~\ref{tab:3dpw}. We observe that our method shows the least performance drop from the testing set to the subset, which demonstrates our method is indeed more robust to inter-person occlusion. 


\begin{table}
	\footnotesize
	\centering
	\begin{tabular}{c|c|c|c}
		\cline{1-4}
		\cline{1-4}
		\rule{0pt}{2.6ex}
		\textbf{Group} & \textbf{Method} & PCK & PCK$_{abs}$\\
		\cline{1-4}
		\rule{0pt}{2.6ex}
		& Mehta et al. \cite{mehta2018single} & 65.0 & n/a\\
		Person- & Rogez et al., \cite{rogez2019lcr} & 70.6 & n/a\\
		centric & Cheng et al. \cite{cheng2019occlusion} & 74.6 & n/a\\
		& Cheng et al. \cite{cheng2020sptaiotemporal} & 80.5 & n/a\\
		\cline{1-4}
		& Moon et al. \cite{Moon_2019_ICCV_3DMPPE} & 82.5 & 31.8\\
		
		Camera- & Lin et al. \cite{lin2020hdnet} & 83.7 & 35.2\\
		centric & Zhen et al. \cite{zhen2020smap} & 80.5 & 38.7\\
		& Li et al. \cite{li2020hmor} & 82.0 & 43.8\\
		& Cheng et al. \cite{cheng2021graph} & \underline{87.5} & \underline{45.7}\\ 
		& Our method & \textbf{89.6} & \textbf{48.0}\\
		\cline{1-4}
		\cline{1-4}
	\end{tabular}
	\vspace{0.5em}
	\caption{Quantitative evaluation on multi-person 3D dataset, MuPoTS-3D. Best in \textbf{bold}, second best \underline{underlined}. }
	\vspace{-0.5em}
	\label{tab:MuPoTS_3d}
\end{table}


\begin{table}
	\footnotesize
	\centering
	\begin{tabular}{c|c|c|c}
		\cline{1-4}
		\rule{0pt}{2.6ex}
		\textbf{Method} & $t=0.4m$ & $t=0.8m$ & $t=1.2m$ \\
		\cline{1-4}
		\rule{0pt}{2.6ex}
		\cite{redmon2018yolov3} + \cite{martinez2017simple} + \cite{rogez2019lcr} & 39.14 & 47.38 & 49.03\\
		LoCO \cite{fabbri2020compressed} & \underline{50.82} & \underline{64.76} & \underline{70.44}\\
		Ours & \textbf{57.22} & \textbf{68.51} & \textbf{72.86}\\
		\cline{1-4}
	\end{tabular}
	\vspace{0.5em}
	\caption{Quantitative results on JTA dataset. F1 values are reported based on different threshold $t$ when the point is considered "true positive" when the distance from corresponding distance is less than $t$. Best in \textbf{bold}, second best \underline{underlined}.}
    \vspace{-0.5em}
	\label{tab:jta}
\end{table}


\begin{table}
	\footnotesize
	\centering
	\begin{tabular}{c|c|c|c}
		\cline{1-4}
		\rule{0pt}{2.6ex}
		\textbf{Group} & \textbf{Method} & MPJPE & PA-MPJPE\\
		\cline{1-4}
		\rule{0pt}{2.6ex}
		& Hossain et al., \cite{hossain2018exploiting} & 51.9 & 42.0 \\
		& Wandt et al., \cite{wandt2019repnet}* & 50.9 & 38.2 \\
		\small{Person-} & Pavllo et al., \cite{pavllo20193d} & 46.8 & 36.5 \\
		\small{centric} & Cheng et al., \cite{cheng2019occlusion} & 42.9 & 32.8\\
		& Kocabas et al., \cite{kocabas2020vibe} & 65.6 & 41.4  \\
		& Kolotouros et al. \cite{kolotouros2019learning} & n/a & \underline{41.1} \\
		\cline{1-4}
		& Moon et al., \cite{Moon_2019_ICCV_3DMPPE} & 54.4 & 35.2 \\
		\small{Camera-} & Zhen et al., \cite{zhen2020smap} & 54.1 & n/a \\
		\small{centric} & Li et al., \cite{li2020hmor} & 48.6 & \underline{30.5} \\
		& Ours & \textbf{40.7} & \textbf{30.4} \\
		\cline{1-4}
	\end{tabular}
	\vspace{0.5em}
	\caption{Quantitative evaluation on Human3.6M for normalized and camera-centric 3D human pose estimation. * denotes ground-truth 2D labels are used. Best in \textbf{bold}, second best \underline{underlined}. }
	\label{tab:h36m}
	\vspace{-0.5em}
\end{table}


\begin{table}
	\footnotesize
	\centering
	\begin{tabular}{c|c|c|c}
		\cline{1-4}
		\rule{0pt}{2.6ex}
		\textbf{Dataset} & \textbf{Method} & PA-MPJPE & $\delta$ \\
		\cline{1-4}
		\rule{0pt}{2.6ex}
		& Doersch et al. \cite{doersch2019sim2real}  & 74.7 & n/a \\
		& Kanazawa et al. \cite{humanMotionKanazawa19} & 72.6 & n/a \\
		Original & Arnab et al. \cite{arnab2019exploiting} & 72.2 & n/a \\
		& Cheng et al. \cite{cheng2020sptaiotemporal} & 71.8 & n/a \\
		& Sun et al. \cite{sun2019human} & 69.5 & n/a \\
		& Kolotouros et al. \cite{kolotouros2019learning}* & \underline{59.2} & n/a \\
		& Kocabas et al., \cite{kocabas2020vibe}* & \textbf{51.9} & n/a \\
		& Our method  & 62.9 & n/a \\
		\cline{1-4}
		& Cheng et al. \cite{cheng2020sptaiotemporal} & 92.3 & +20.5\\
		& Sun et al. \cite{sun2019human} & 84.4 & \underline{+14.9}\\
		Subset & Kolotouros et al. \cite{kolotouros2019learning}*  & 79.1 & +19.9\\
		& Kocabas et al., \cite{kocabas2020vibe}* & \textbf{72.2} & +20.3 \\
		& Our method  & \underline{75.6} & \textbf{+12.7}\\
		\cline{1-4}
	\end{tabular}
	\vspace{0.5em}
	\caption{Quantitative evaluation using PA-MPJPE on original 3DPW test set and its occlusion subset. * denotes extra 3D datasets were used in training. Best in \textbf{bold}, second best \underline{underlined}.}
	\vspace{-1em}
	\label{tab:3dpw}
\end{table}

\begin{figure*}[t]
	\centering
	\makebox[\textwidth]{\includegraphics[width=\textwidth]{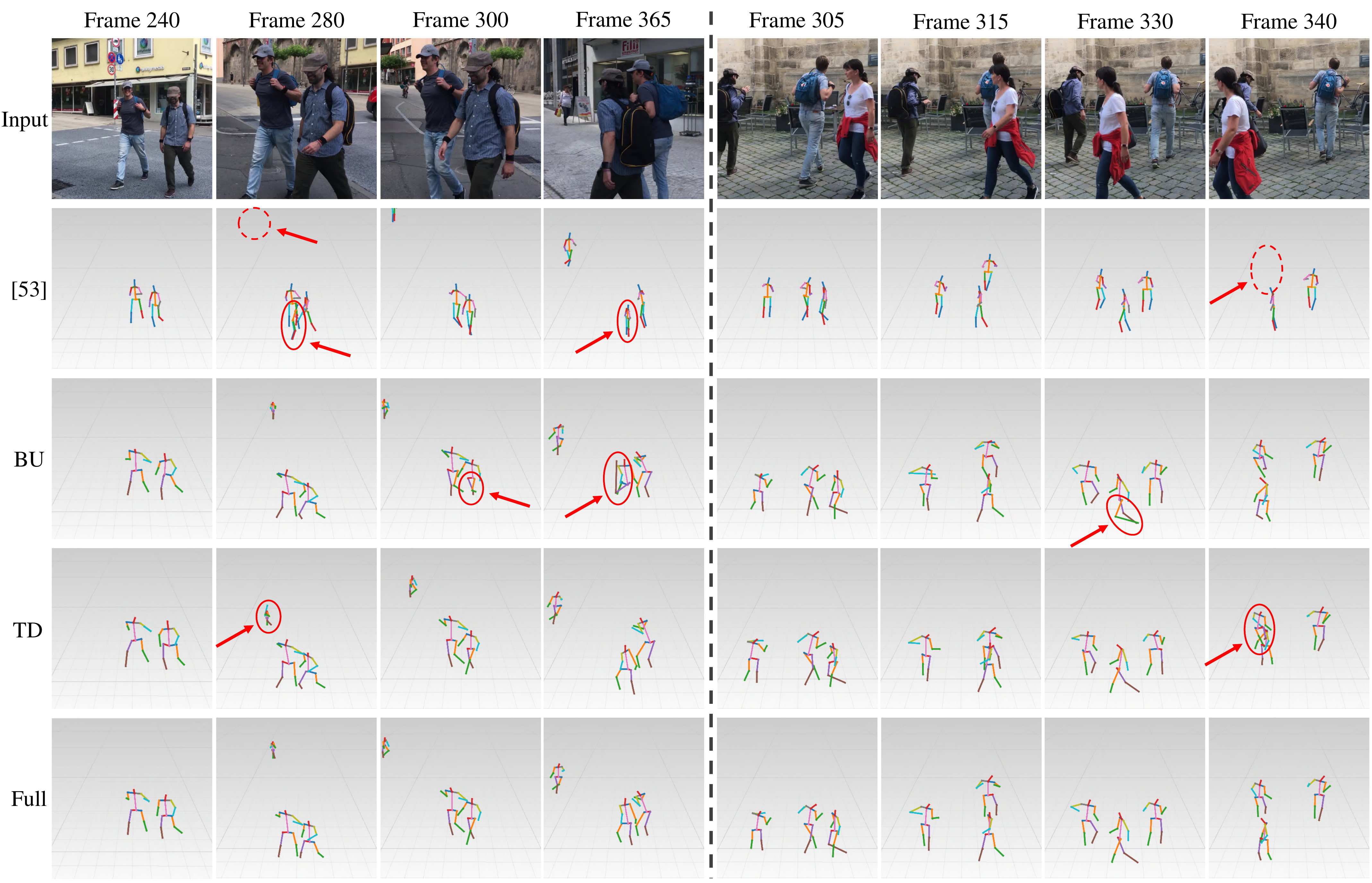}}
	\caption{Examples of results from our whole framework compared with different baseline results. First row shows the images from two video clips; second row shows the results from SMAP \cite{zhen2020smap}; third row shows the result of of our bottom-up (BU) branch; fourth row shows the results of our top-down (TD) branch; last row shows the results of our full model. Wrong estimations are labeled with red circles.
	}
	\label{fig:qualitative_evaluation}
	\vspace{-1em}
\end{figure*}

\vspace{-1.0em}
\paragraph{Qualitative Evaluation}

Fig. \ref{fig:qualitative_evaluation} shows the comparison among a SOTA bottom-up method SMAP \cite{zhen2020smap}, our bottom-up branch, top-down branch, and full model. 
We observe that SMAP suffers from person scale variation where the person who is far from the camera is missing in frame 280 as well as inter-occlusion (e.g., frame 365 and 340). 
Our bottom-up branch is robust to scale variance, but fragile to the out-of-image poses as our discriminator is not used here (e.g., frame 365 and 330). 
Moreover, our top-down branch produces reasonable relative poses with the aid of GCN and TCNs. 
However, there exists error of camera-centric root depth in our top-down branch, because our top-down branch estimates root depth based on individual 2D poses and lacks global awareness (e.g., frame 280). 
Finally, our full model benefits from both branches and produces the best 3D pose estimations among these baselines. 

\begin{figure}
    \centering
    \includegraphics[width=\linewidth]{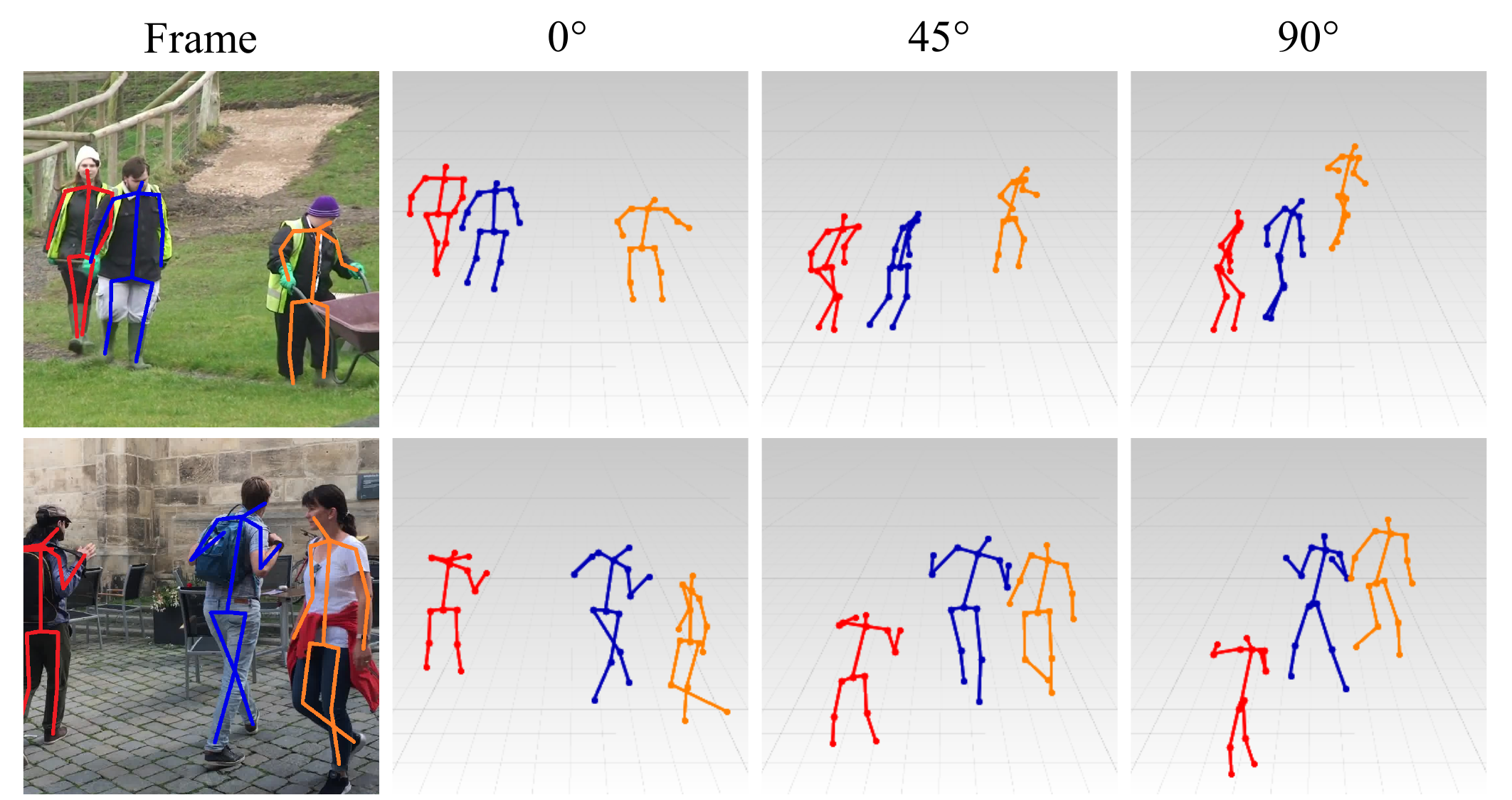}
    \caption{Qualitative results of the estimated 2D poses overlaying on input images and the estimated 3D poses visualized in novel viewpoints (virtual camera rotated by 0, 45, 90 degrees clockwise).
    Different colors are used for different persons in both 2D and 3D human poses for better visualization purpose.}
    \label{fig:multi_persp}
    \vspace{-0.8em}
\end{figure}

\begin{figure}
    \centering
    \includegraphics[width=\linewidth]{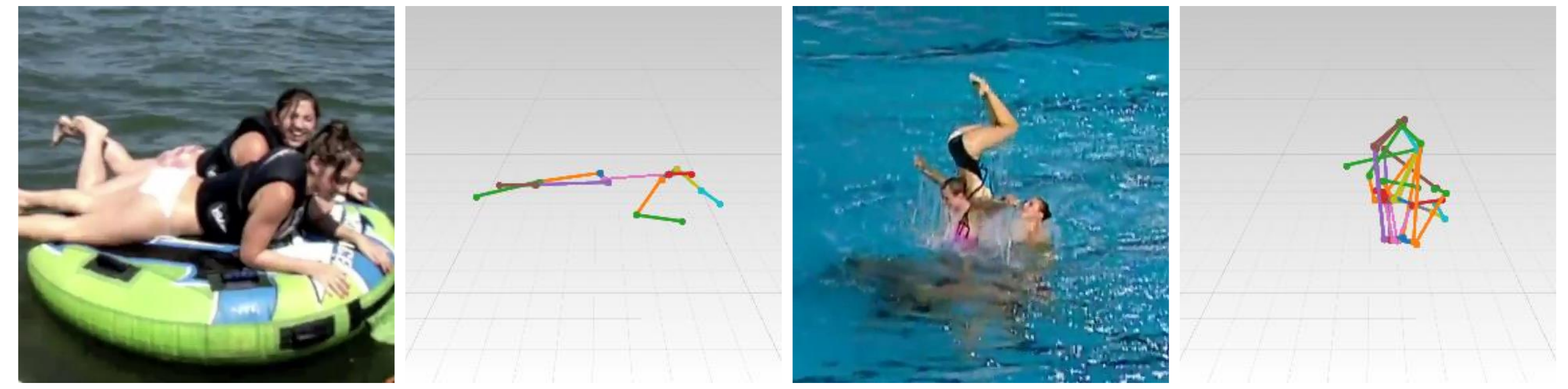}
    \caption{Two representative failure cases of our method. 
    }
    \label{fig:failure}
    \vspace{-0.8em}
\end{figure}

We also provide results of the estimated 3D poses in novel viewpoints and the estimated 2D poses overlaid on input images as in Fig.~\ref{fig:multi_persp} where our estimated camera-centric 3D poses visualized from different angles further validate the effectiveness of our method. 
Two failure cases are shown in Fig.~\ref{fig:failure} where the samples are taken from MPII dataset. The common failure cases are constant heavy occlusion (left) and unusual poses (right).

\vspace{-0.5em}
\section{Conclusion}
\vspace{-0.3em}
We have proposed a novel method for  monocular-video 3D multi-person pose estimation, which addresses the problems of   inter-person occlusion and close interactions.
We introduced the integration of top-down and bottom-up approaches to exploit their strengths. 
%
%
Our quantitative and qualitative evaluations show the effectiveness of our method compared to the state-of-the-art baselines.

\vspace{-0.5em}
\section*{Acknowledgements}
\vspace{-0.3em}
This research is supported by the National Research Foundation, Singapore under its Strategic Capability Research Centres Funding Initiative. Any opinions, findings and conclusions or recommendations expressed in this material are those of the author(s) and do not reflect the views of National Research Foundation, Singapore.

{\small
\bibliographystyle{ieee_fullname}
\bibliography{egbib}
}

\clearpage
\setcounter{section}{0}

\subfile{supp.tex}

\end{document}

%% file: supp.tex



\begin{center}
\Large\bfseries Supplementary Material
\end{center}

\section{Network Structure}


\paragraph{GCN Structure} Unlike existing GCN methods which use an undirected graph \cite{ci2019optimizing,cai2019exploiting}, we use a directed graph. The advantage of using directed graph is that more reliable joints with higher confidence are capable to influence the unreliable ones with low confidence with non-symmetric adjacency matrix. We adopt the GCN method following~\cite{cheng2021graph}.

The features are propagated according to an adjacent matrix in GCNs, implying the edge values in the propagation graph.
Given the heatmap $H$ from the 2D pose estimator, we choose the location of the highest value in the map as a vertex in the graph for each joint, and the adjacency matrix is formed by the following equation:
\begin{equation}
\mathbb{A}_{i,j} = 
\begin{cases}
\max(H_i) \exp(-order(i,j)) &(i\ne j)\\
\max(H_i) &(i=j)\\
\end{cases},
\label{eq:adj}
\end{equation}
where the $A_{i,j}$ is the outward weight from vertex $i$ to vertex $j$. $max(H_i)$ stands for the confidence of the $i_{th}$ joint. $order(i,j)$ is the minimal number of hops that is required to reach vertex $j$ from vertex $i$. 
This formation of adjacency imposes more weight for close vertices and less for far ones. More details please refer to~\cite{cheng2021graph}.

\begin{figure}[]
    \centering
    \includegraphics[width=\linewidth]{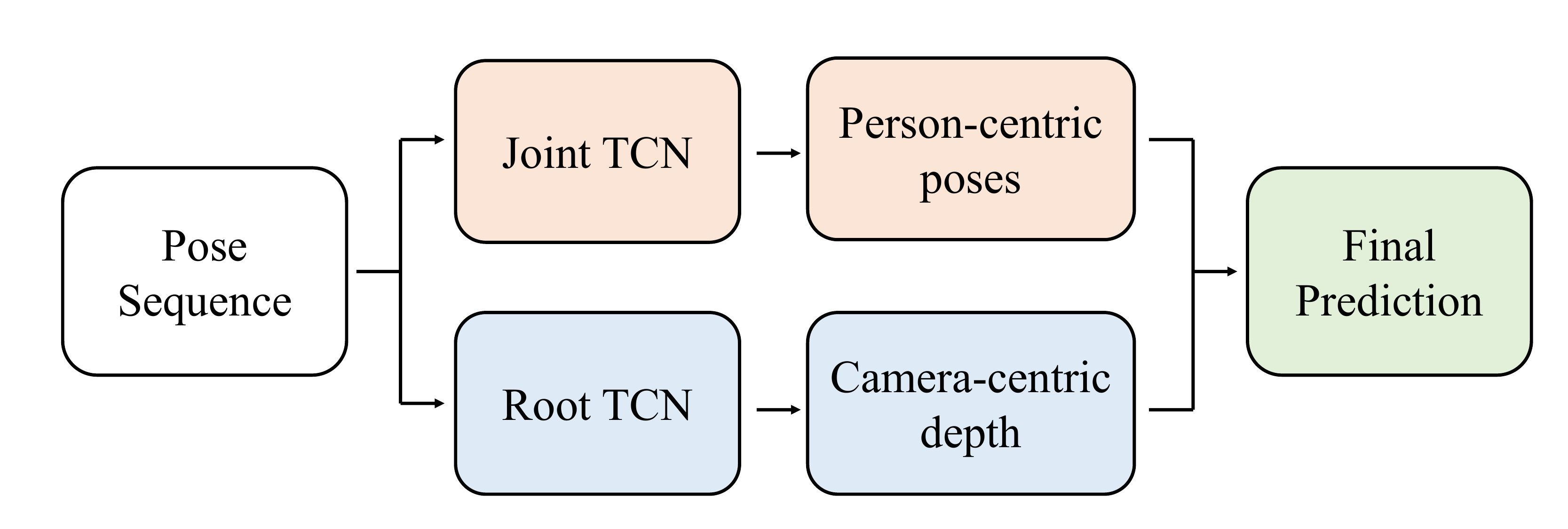}
    \caption{Pipeline of our TCNs. Our TCNs include one Joint TCN for relative pose estimation and one Root TCN for camera-centric root depth estimation. }
    \label{fig:tcnpipeline}
\end{figure}

\paragraph{TCN Structure}
Our GCN can complete the pose under occlusion or missing information, yet produces jittering results because of its lack of temporal smoothness. Previous works on the Temporal Convolutional Network (TCN) show the effectiveness of a TCN to constrain the temporal smoothness of predicted 3D poses~\cite{pavllo20193d,cheng2019occlusion}. We adopt the TCN structure~\cite{pavllo20193d}. As shown in Fig. \ref{fig:tcnpipeline}, we utilize two TCNs to estimate the person-centric 3D poses (i.e., joints) and the camera-centric root joint depths, respectively. We named the two TCNs as: Joint-TCN and Root-TCN.

The Joint-TCN takes the 3D pose sequence produced by our GCN as input, and outputs the refined person-centric 3D poses by considering the temporal information. The loss is L2 between the estimated pose $P^{TCN}$ and its ground-truth $\Tilde{P}$, formulated as:
\begin{equation}
    L_{JTCN} = \frac{1}{K} \sum^{K}_{k=0} |P^{TCN}_k - \Tilde{P}_k|^2,
    \label{eq:JTCN}
\end{equation}
where $K$ is the number of the  joints. 

The Root-TCN takes the 3D pose sequence generated by the GCN and the 2D pose sequence produced by the pose estimator as input, and outputs the estimated camera-centric root depths. 
Instead of directly estimating the camera-centric depth $Z$, we estimate the normalized root depth, which is $R^{TCN} = \frac{Z}{f}$ based on focal length $f$ to avoid the influence of the camera intrinsic parameters. The loss function is L2 between the estimated $R^{TCN}$ and its ground truth $\Tilde{R}$:
\begin{equation}
    L_{RTCN} = \frac{1}{K} \sum^{K}_{k=0} |R^{TCN}_k - \Tilde{R}_k|^2
    \label{eq:RTCN}
\end{equation}
where $K$ is the number of the joints. Based on the person-centric 3D pose from Eq.~(\ref{eq:JTCN}) and the root-joint depth from Eq.~(\ref{eq:RTCN}), the camera-centric 3D pose is obtained. 

\begin{figure}
    \centering
    \includegraphics[width=0.8\linewidth]{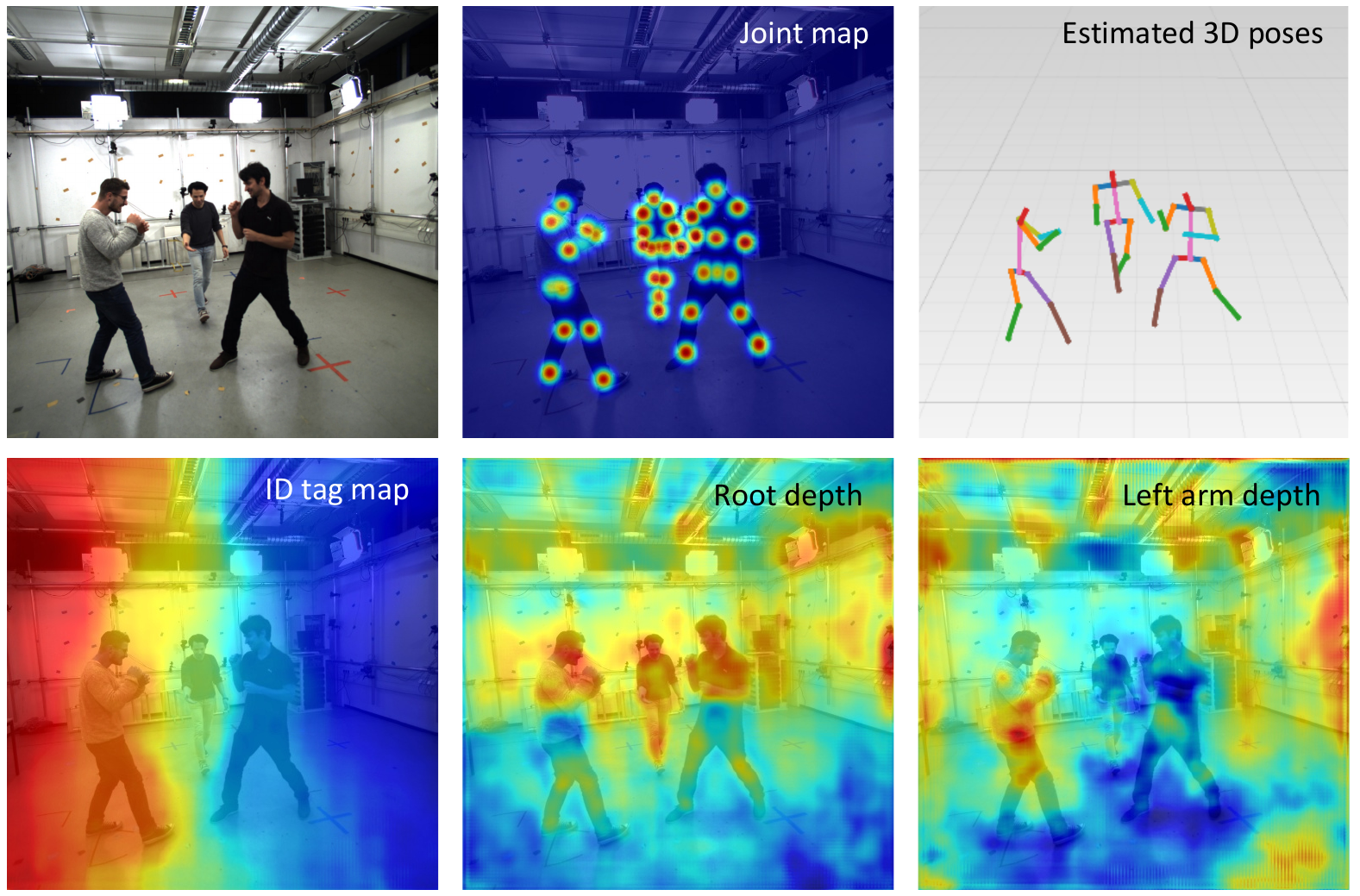}
    \caption{Visualization of estimated heatmaps from the bottom-up branch.}
    \label{fig:heatmaps}
\end{figure}

\paragraph{Illustration of the heatmaps estimated from the bottom-up network} Fig.~\ref{fig:heatmaps} illustrates an example output of the four heatmaps estimated by our bottom-up network. Top left is an input image. Top middle is a joint map, which shows the heatmap of joints where all channels are merged together for better visualization of all joints. Top right is the estimated 3D poses. Bottom left shows the ID tag distribution. Bottom middle is the root depth map where the red represents a person is farther to camera than others. Bottom right is an example of relative depth map with respect to pelvis joint, where left arm depth is used as an example. The arm of left person is farther from the camera (red) compared to his pelvis while the right person's is closer to camera (blue) with respect to his pelvis.

\paragraph{Details of Semi-Supervised Learning}

\begin{figure}
    \centering
    \includegraphics[width=\linewidth]{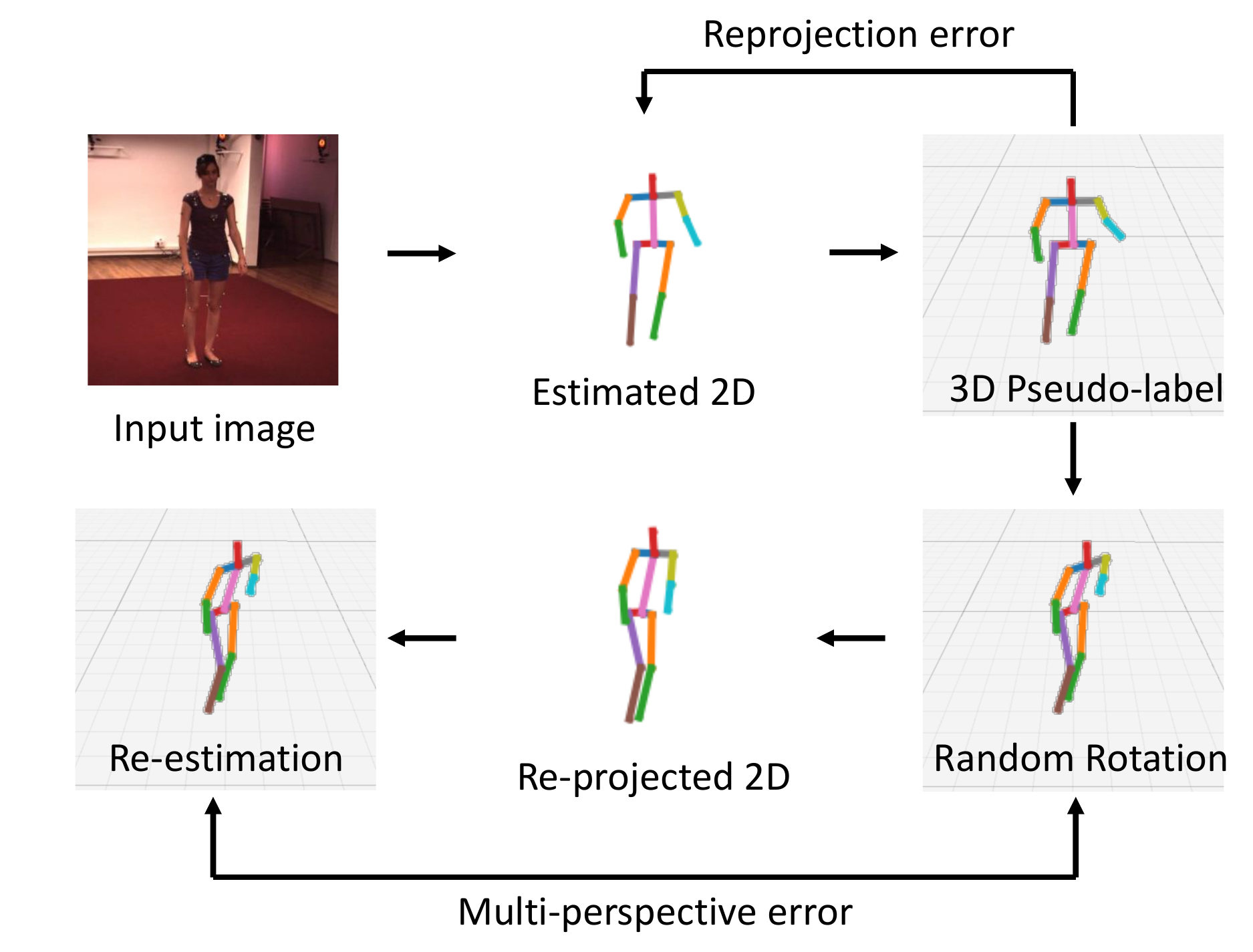}
    \caption{The illustration of our SSL pipeline. The SSL aims to keep two consistency: reprojection and multi-perspective.}
    \label{fig:sslpipeline}
\end{figure}

Our Semi-supervised Learning (SSL) pipeline is shown in Fig. \ref{fig:sslpipeline}. First, we use the trained model to generate the pseudo-label of the unlabelled data, which is the COCO dataset in our experiment. Note that, we use only the images, and not the 2D ground-truths of the joints to mimic the unlabelled data scenario. 
Unfortunately, the pseudo-labels cannot be directly used because some of them are incorrect. Therefore, we use two consistency terms to measure the quality of all the pseudo-labels: the reprojection error and multi-perspective error as mentioned in the main paper. 

As the pose variations of 2D datasets are more abundant than those of 3D datasets, e.g. COCO compared to H36M, the estimated 2D poses are more robust than the estimated 3D poses in terms of different environments and poses. Existing reprojection error \cite{wandt2019repnet} measures the deviation between generated 3D poses and detected 2D poses. Unlike this, we make use of the confidence of the joints from the 2D pose heatmap as weight in computating  the reprojection error to adjust adaptively how much we should enforce the reprojected 3D poses to match the estimated 2D poses based on the confidence of the joints. Thus, the reprojection error  is formulated as:
\begin{equation}
    E_{rep} = \frac{1}{K} \sum^{K}_{k=1} C_k |rep(X_{3D,k}) - X_{2D,k}|^2
\end{equation}
where the $X_{3D}$ is the predicted 3D pose from the network, and $X_{2D}$ stands for the 2D estimations from our multi-person 2D pose estimator. $rep(\cdot)$ is the reprojection function from 3D to 2D. $K$ stands for the number of joints in total. Moreover, the error is a weighted sum of each joint's confidence score $C_k$, which is explained in Eq.~(\ref{eq:adj}).

We follow \cite{chen2019unsupervised} to use a multi-perspective error as an additional measure to enforce the consistency of the predicted 3D poses from different viewing angles. Given a pseudo-label 3D pose $P^{pse}_{3D}$, we randomly rotate the pose along $y$ axis (i.e., y-axis is perpendicular to the ground plane) to obtain $P'^{pse}_{3D}$, and re-project it to the 2D coordinates $P'^{pse}_{2D}$. Finally, we predict the $P''^{pse}_{3D}$ based on the re-projection. 

\section{Implementation Details}


\paragraph{Multi-Person Pose Estimator}

Our multi-person pose estimator uses HRNet-w32 \cite{sun2019hrnet} as the backbone and is trained on the combination of the MuCO and COCO dataset. We duplicate the COCO dataset twice to balance the training data between two datasets. The network is trained with the Adam optimizer with learning rate starts at $0.001$ and decreases to $\frac{1}{10}$ at epoch $30$ and $40$. The network is trained for 50 epochs and it takes 35 hours to train on 8x RTX Quadro 8000 GPUs. 

\paragraph{GCN and TCNs}

Our GCN and TCNs are trained based on the pre-extracted heatmaps from our multi-person pose estimator. We train the networks with the Adam optimizer with learning rate starts at $0.001$ and decrease to $\frac{1}{10}$ every 40 epochs. The networks are trained with 100 epochs and takes 25 hours on single RTX 2080Ti GPU. We use the augmentation mentioned in \cite{cheng2019occlusion} to train the network to better handle the occlusion.

\paragraph{Bottom-Up Network}

Our bottom-up network is trained based on the combination of the MuCO and COCO dataset. To balance the number of training samples, we duplicate the COCO dataset twice and combine with the MuCO dataset. The bottom-up network is trained with the Adam optimizer with learning rate starts at $0.001$ and decrease to $\frac{1}{10}$ at the $30^{th}$ and $40^{th}$ epoch. The network is trained for 50 epochs and it takes 65 hours on 8x RTX Quadro 8000 GPUs.

\paragraph{Integration Network}

Our integration network contains $5$ fully connected layers with layer size $512$. The network is trained with the Adam optimizer with learning rate $0.001$ in beginning, and decreased to $\frac{1}{10}$ every 50 epochs. The network is trained for $150$ epochs and takes 3.5 hours on single RTX 2080Ti GPU. 
The data augmentation procedure is discussed in the main paper. We briefly explain here for clarity: 1) We use random masking to simulate the occlusion, where the occluded joints are masked to $(0,0)$. 2) We apply a random shifting of joints based on a Gaussian random to simulate the inaccurate pose estimation. 3) We randomly make one of the poses in the pair to be zero, to simulate the unpaired poses. 

\section{Datasets Description} 
\label{datasets}

\paragraph{MuPoTS-3D}~\cite{mehta2018single} is a 3D multi-person testing set that consists of $>$8000 frames of 5 indoor and 15 outdoor scenes, and its corresponding training set is augmented from 3DHP, called MuCo-3DHP. The ground-truth 3D pose of each person in a video is obtained from multi-view markerless motion capture system, which is suitable for evaluating 3D multi-person pose estimation performance in both person-centric and camera-centric coordinates.
Following~\cite{Moon_2019_ICCV_3DMPPE}, the training set (MuCo-3DHP) is used for training our bottom-up network, and MuPoTS-3D is used only for performance evaluation.

\paragraph{JTA}~\cite{fabbri2018learning} is a synthesized dataset from Grand Theft Auto V (GTA-V) game scene including various of illumination, viewpoints, and occlusion. It is a multi-person dataset with at most $32$ persons appear in one frame. In addition, the images also demonstrate large person size variation as the crowd spread from close to camara and far from camera in various scenes. Because of these reasons, even it is a synthetic dataset, we'd like to perform evaluation on it. The dataset contains 512 videos, in which there are 256, 128, 128 for training, validation and testing, respectively. We follow the work \cite{fabbri2020compressed} to estimate the F1 score under different distance threshold as a performance evaluation metric. 

\vspace{-0.5em}
\paragraph{Human3.6M}~\cite{human36ionescu} is widely used for 3D human pose estimation. The dataset contains 3.6 million single-person images where an actor performs different activities in mocap studio at each video clip, so it is suitable for evaluation of 3D single-person pose estimation. 
Human3.6M is used for evaluating person-centric pose estimation performance. Following previous works~\cite{hossain2018exploiting,pavllo20193d,wandt2019repnet}, the subject 1,5,6,7,8 are used for training, and 9 and 11 for testing.

\vspace{-0.5em}
\paragraph{3DPW}~\cite{3DPW} is an outdoor multi-person video dataset for 3D human pose reconstruction.
%
In each video, one target person wearing inertial measurement units (IMUs) performs daily activities outdoor, so 3D ground-truth is available for the target person only. 
%
Following previous methods~\cite{humanMotionKanazawa19,sun2019human}, we use 3DPW for testing without any fine-tuning. 
The ground-truth of 3DPW is SMPL 3D mesh model~\cite{loper2015smpl}, where the definition of joints differs from what is used in 3D human pose estimation (skeleton-based) like Human3.6M, so 3DPW is rarely used in the evaluation of skeleton-based methods~\cite{tripathi2020posenet3d}.

\begin{figure}[t]
    \centering
    \includegraphics[width=\linewidth]{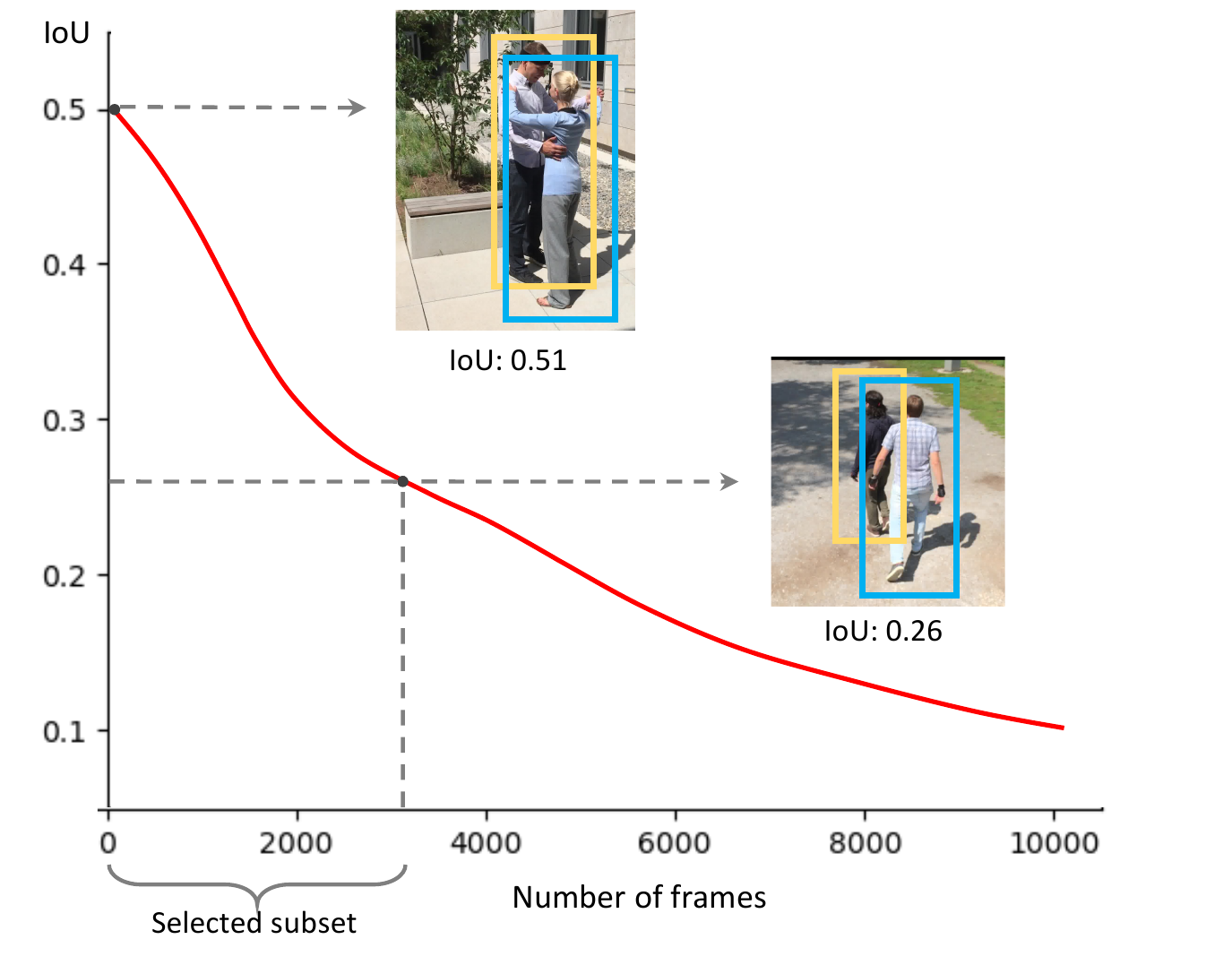}
    \caption{Interaction IoUs of 3DPW test set. 
    }
    \label{fig:iou}
\end{figure}

\begin{figure}[t]
    \centering
    \includegraphics[width=\linewidth]{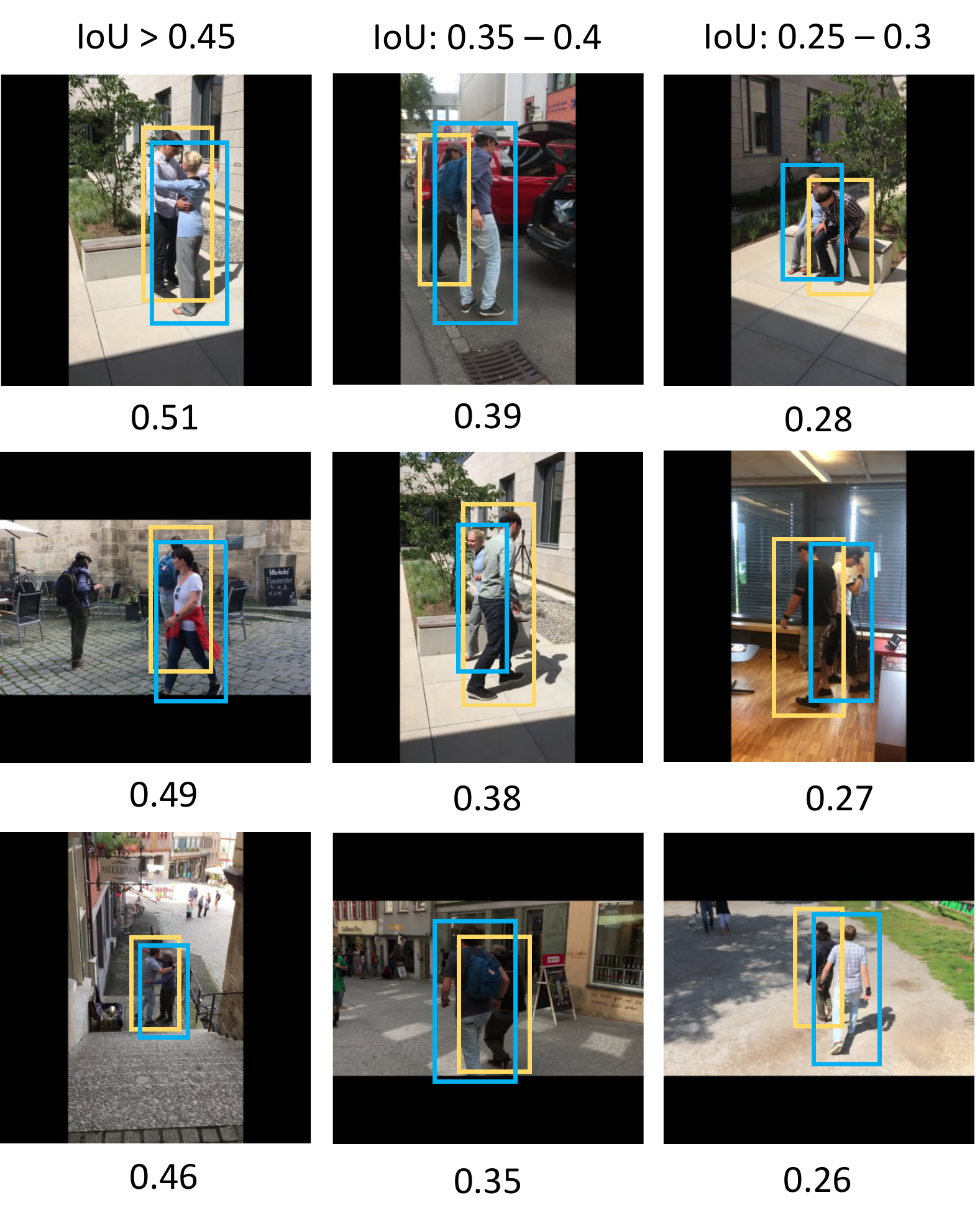}
    \caption{Some sample images of different IoU level that are selected for inter-person occlusion subset. IoU values are added below each image.}
    \label{fig:iousample}
\end{figure}

Evaluation errors on 3DPW cannot objectively reflect the performance of the skeleton-based methods, due to different definitions of joints. We select the top 3000 frames with the largest IoU between the target person (i.e., the person with 3D ground-truth label) and other persons based on detection out of 3DPW test set to create an inter-person occlusion subset, and then perform evaluation on it. The IoU statistics of the 3DPW test set is shown in Fig. \ref{fig:iou}, and the threshold at $3000^{th}$ frame is $0.26$. Some samples of different occlusion level is shown in Fig. \ref{fig:iousample}. 

In fact, the error on this subset is still not a good performance indicator, the performance change of a method between the full testing set and this subset can measure how well the method can handle the inter-person occlusion problem. As shown in Table 6 in the main paper, our method shows the smallest error increase among all the existing methods, which demonstrates that our method is indeed capable of handling inter-person occlusion more effectively.


\paragraph{Training Datasets}


Both the 2D datasets and 3D datasets are used to train our networks. In the following, we explain the details of the used datasets in the training processes of our pose estimator, top-down and bottom-up networks, pose discriminator, and semi-supervised learning. 
\begin{itemize}
	\item 
	\textit 2D datasets for pose estimator training: We use both COCO and MuCO for training the multi-person pose estimator. Because the MuCO dataset is a synthesized dataset, solely training on the MuCO dataset will result in overfitting problem and produces unstable predictions on natural or wild images. Therefore, COCO is included for enhance the generalization ability of the network. 
	
	\item 
	3D dataset for top-down network training: We use MuCO and its original 3DHP dataset to train the GCN and TCNs in the top-down network. MuCO and 3DHP are used for the GCN on single frame pose refinement, while the 3DHP is used to train the TCN that incorporates the temporal information. Since the network works on the $x,y,z$ coordinates, no overfitting problem was observed from the trained models.

\item 
3D dataset for bottom-up network training: We use both MuCO and COCO to train the bottom-up network. We additionally include COCO, which is used only for training joint heatmaps and ID tag maps. 

\item
3D dataset for pose discriminator training: MuCO is used for training the integration net and pose discriminator. In addition, we do random translation and rotation of the poses to generate more synthesized interaction pairs.

\item
Additional 2D data for semi-supervised learning training: We use COCO for the unlabeled image dataset in training our semi-supervised learning.	
\end{itemize}

\paragraph{Evaluation Protocols}

While we include the discussion of the datasets for the tables in the main paper, here we provide the details for the sake of clarity. 
Our model is trained with the datasets explained in the previous section (i.e., Training Datasets Used) for ablation study in Table 1 and 2, evaluations in Table 3 (MuPoTS-3D),  Table 5 (Human3.6M), and Table 6 (3DPW). 
 
The JTA dataset is captured from computer game, which has a domain gap to the real-world images. To perform the evaluation on the JTA dataset in Table 4 (JTA), we use the JTA training set to re-train the whole pipeline and perform the evaluation on the JTA test set. 

As mentioned in the 3DPW dataset explanation, we follow the literature \cite{humanMotionKanazawa19,sun2019human} and only perform testing on 3DPW. 
%
%
Note that, the SOTA methods \cite{kolotouros2019learning,kocabas2020vibe} both use additional 2D and 3D datasets in training their networks. We do not use the 3DPW dataset to train our network, but used it to train the joint adaptation network \cite{tripathi2020posenet3d}, which transfers our predicted 3D poses of MuCO joint's definition to that of 3DPW defined on the SMPL model \cite{loper2015smpl}. 



\section{Detailed Experimental Results}

As our method focuses on the 3D multi-person scenarios, our network is trained on the 3D multi-person datasets as discussed in section \ref{datasets}. To have a fair comparison against existing methods that are trained only with the single-person Human3.6M dataset, we re-trained the whole pipeline from scratch on H3.6M dataset following the training protocols~\cite{hossain2018exploiting,pavllo20193d}. 
%
%
The evaluation result on the person-centric 3D human pose estimation is shown in Table~\ref{tab:h36msupp}. Similar to Table 5 in the main paper, our method achieves comparable performance against the SOTA top-down or bottom-up 3D multi-person pose estimation methods \cite{Moon_2019_ICCV_3DMPPE,zhen2020smap,li2020hmor} on this single-person dataset. 

Following \cite{Moon_2019_ICCV_3DMPPE}, we also calculate our method's accuracy using the MPRE metrics, which measures the camera-centric 3D human pose estimation performance. In particular, \cite{Moon_2019_ICCV_3DMPPE} is 120.0, ours is 86.5, which shows $27.9\%$ error reduction. HDNet \cite{lin2020hdnet} reports a better value on MPRE as 77.6, however, their method can only handle single-person cases, and performs poorly on multi-person cases where their value of PCK$_{abs}$ is 35.2, but ours is 48.0, which is a $36.4\%$ improvement. As camera-centric 3D pose estimation is for multi-person scenario, showing good result on single-person dataset but poor performance on multi-person dataset is not applicable to the real problem in 3D multi-person pose estimation. 

\begin{table}[t]
	\footnotesize
	\centering
	\begin{tabular}{c|c|c}
		\cline{1-3}
		\rule{0pt}{2.6ex}
		\textbf{Method} & MPJPE & PA-MPJPE\\
		\cline{1-3}
		Moon et al., \cite{Moon_2019_ICCV_3DMPPE} & 54.4 & 35.2 \\
		Zhen et al., \cite{zhen2020smap} & 54.1 & n/a \\
		Li et al., \cite{li2020hmor} & \underline{48.6} & \textbf{30.5} \\
		Ours & \textbf{42.1} & \underline{31.6}  \\
		\cline{1-3}
	\end{tabular}
	\vspace{0.5em}
	\caption{Quantitative evaluation on Human3.6M for person-centric 3D human pose estimation. Best in \textbf{bold}, second best \underline{underlined}. }
	\label{tab:h36msupp}
	\vspace{-1.0em}
\end{table}

\begin{table*}
    \centering
    \begin{tabular}{cccccccccccc}
        \noalign{\hrule height 1.5pt}
         Method & S1 & S2 & S3 & S4 & S5 & S6 & S7 & S8 & S9 & S10 & - \\ \hline
         Moon et al. \cite{Moon_2019_ICCV_3DMPPE} & 59.5 & 44.7 & 51.4 & 46.0 & 52.2 & 27.4 & 23.7 & 26.4 & 39.1 & 23.6 & \\
         Zhen et al. \cite{zhen2020smap} & 41.6 & 33.4 & 45.6 & 16.2 & 48.8 & 25.8 & \textbf{46.5} & 13.4 & 36.7 & \textbf{73.5} & \\
         Ours & \textbf{69.2} & \textbf{57.1} & \textbf{49.3} & \textbf{68.9} & \textbf{55.1} & \textbf{36.1} & 49.4 & \textbf{33.0} & \textbf{43.5} & 52.8 & \\\noalign{\hrule height 1.5pt}
         Method & S11 & S12 & S13 & S14 & S15 & S16 & S17 & S18 & S19 & S20 & Avg \\ \hline
         Moon et al. \cite{Moon_2019_ICCV_3DMPPE} & 18.3 & 14.9 & 38.2 & 26.5 & 36.8 & 23.4 & 14.4 & 19.7 & 18.8 & 25.1 & 31.5 \\
         Zhen et al. \cite{zhen2020smap} & \textbf{43.6} & 22.7 & 21.9 & 26.7 & 47.1 & 32.5 & 31.4 & 18.0 & 33.8 & 47.8 & 35.4 \\
         Ours & 48.8 & \textbf{36.5} & \textbf{51.2} & \textbf{37.1} & \textbf{47.3} & \textbf{52.0} & \textbf{20.3} & \textbf{43.7} & \textbf{57.5} & \textbf{50.4} & \textbf{48.0} \\\noalign{\hrule height 1.5pt}
    \end{tabular}
    \vspace{0.5em}
    \caption{$PCK_{abs}$ on MuPoTS-3D dataset for all poses. Best in \textbf{bold}.}
    \label{tab:pckabs}
\end{table*}

\begin{table*}
    \centering
    \begin{tabular}{cccccccccccc}
        \noalign{\hrule height 1.5pt}
         Method & S1 & S2 & S3 & S4 & S5 & S6 & S7 & S8 & S9 & S10 & - \\ \hline
         Rogez et al. \cite{rogez2017lcr} & 67.7 & 49.8 & 53.4 & 59.1 & 67.5 & 22.8 & 43.7 & 49.9 & 31.1 & 78.1 &  \\
         Rogez et al. \cite{rogez2019lcr} & 87.3 & 61.9 & 67.9 & 74.6 & 78.8 & 48.9 & 58.3 & 59.7 & 78.1 & 89.5 & \\
         Dabral et al. \cite{dabral2018learning} & 85.1 & 67.9 & 73.5 & 76.2 & 74.9 & 52.5 & 65.7 & 63.6 & 56.3 & 77.8 & \\
         Mehta et al. \cite{mehta2018single} & 81.0 & 59.9 & 64.4 & 62.8 & 68.0 & 30.3 & 65.0 & 59.2 & 64.1 & 83.9 & \\
         Mehta et al. \cite{mehta2017vnect} & 88.4 & 65.1 & 68.2 & 72.5 & 76.2 & 46.2 & 65.8 & 64.1 & 75.1 & 82.4 & \\
         Zhen et al. \cite{zhen2020smap} & 88.8 & 71.2 & 77.4 & 77.7 & 80.6 & 49.9 & 86.6 & 51.3 & 70.3 & 89.2 & \\
         Moon et al. \cite{Moon_2019_ICCV_3DMPPE} & \textbf{94.4} & 77.5 & 79.0 & 81.9 & 85.3 & 72.8 & 81.9 & 75.7 & \textbf{90.2} & 90.4 & \\
         Ours & 93.4 & \textbf{91.3} & \textbf{84.7} & \textbf{83.3} & \textbf{89.1} & \textbf{85.2} & \textbf{95.4} &  \textbf{92.1} & 89.5 & \textbf{93.1} & \\\noalign{\hrule height 1.5pt}
         Method & S11 & S12 & S13 & S14 & S15 & S16 & S17 & S18 & S19 & S20 & Avg \\ \hline
         Rogez et al. \cite{rogez2017lcr} & 50.2 & 51.0 & 51.6 & 49.3 & 56.2 & 66.5 & 65.2 & 62.9 & 66.1 & 59.1 & 53.8 \\
         Rogez et al. \cite{rogez2019lcr} & 69.2 & 73.8 & 66.2 & 56.0 & 74.1 & 82.1 & 78.1 & 72.6 & 73.1 & 61.0 & 70.6 \\
         Dabral et al. \cite{dabral2018learning} & 76.4 & 70.1 & 65.3 & 51.7 & 69.5 & 87.0 & 82.1 & 80.3 & 78.5 & 70.7 & 71.3 \\
         Mehta et al. \cite{mehta2018single} & 67.2 & 68.3 & 60.6 & 56.5 & 59.9 & 79.4 & 79.6 & 66.1 & 66.3 & 63.5 & 65.0 \\
         Mehta et al. \cite{mehta2017vnect} & 74.1 & 72.4 & 64.4 & 58.8 & 73.7 & 80.4 & 84.3 & 67.2 & 74.3 & 67.8 & 70.4 \\
         Zhen et al. \cite{zhen2020smap} & 72.3 & 81.7 & 63.6 & 44.8 & 79.7 & 86.9 & 81.0 & 75.2 & 73.6 & 67.2 & 73.5 \\
         Moon et al. \cite{Moon_2019_ICCV_3DMPPE} & 79.2 & 79.9 & 75.1 & 72.7 & 81.1 & 89.9 & 89.6 & 81.8 & 81.7 & 76.2 & 81.8 \\
         Ours & \textbf{85.4} & \textbf{85.7} & \textbf{89.9} & \textbf{90.1} & \textbf{88.8} & \textbf{93.7} & \textbf{92.2} & \textbf{87.9} & \textbf{89.7} & \textbf{91.9} & \textbf{89.6} \\\noalign{\hrule height 1.5pt}
    \end{tabular}
    \vspace{0.5em}
    \caption{$PCK$ on MuPoTS-3D dataset for all poses. Best in \textbf{bold}.}
    \label{tab:pckrel}
\end{table*}

To have a better understanding on how our method compare with existing methods for each test sequence in MuPoTS-3D dataset, extended version of Table 3 in the main paper for each test sequence is summarized in Table \ref{tab:pckabs} and \ref{tab:pckrel} for the camera-centric and person-centric evaluations using $PCK_{abs}$ and $PCK$ metrics. We observe that our method consistently outperforms other methods in both the camera-centric and person-centric 3D multi-person pose estimation.

\section{More Qualitative Results}

In this section, we provide additional results compared with the SOTA 3D multi-person pose estimation methods. In the main paper, we already provided a qualitative comparison on 3DPW test set in Fig. 5, where the results of SMAP \cite{zhen2020smap} is used as they released their code and we can perform testing with it. 

\begin{figure*}
    \centering
    \includegraphics[width=\textwidth]{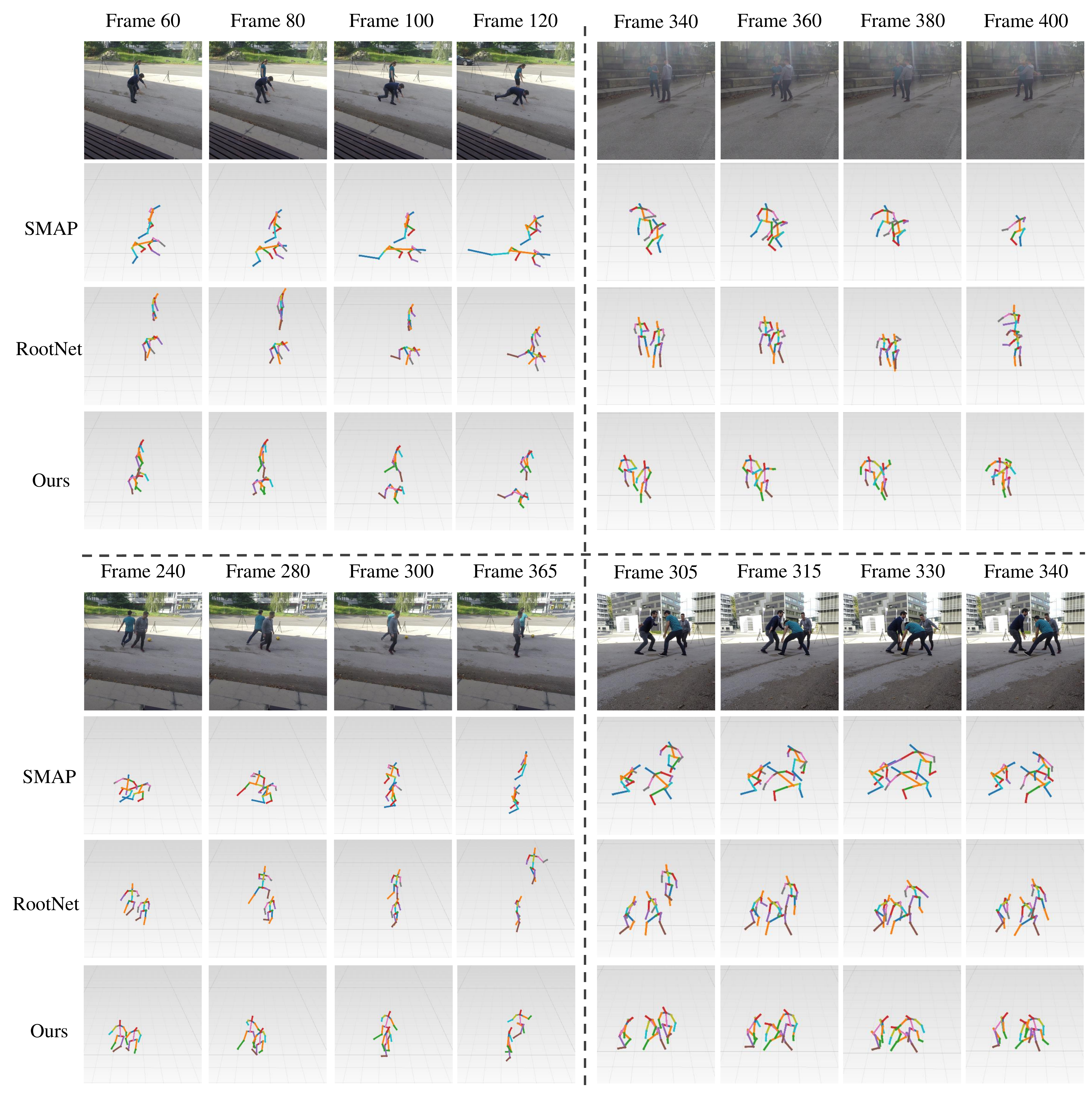}
    \caption{Results of our method compared with that of SMAP \cite{zhen2020smap} (i.e., the SOTA bottom-up method) and RootNet \cite{Moon_2019_ICCV_3DMPPE} (i.e., the SOTA top-down method) on MuPoTS dataset. Results from four video clips are included: top-left, top-right, bottom-left, and bottom-right. For each video clip, the first row is the frames from the video clip; the second row is the result of SMAP; the third row is the result of RootNet; the fourth row is the result of our method. It is observed from these results that the SOTA methods suffer from inter-person occlusions while our method can handle these challenges and produce accurate camera-centric 3D multi-person pose estimation.}
    \label{fig:mupots}
\end{figure*}

\paragraph{Additional Comparison on MuPoTS-3D} 
To compare with more methods, we provide additional results on MuPoTS-3D as RootNet \cite{Moon_2019_ICCV_3DMPPE} released their pretrained model on this dataset, so we can perform testing on MuPoTS-3D using their released model. Together with SMAP \cite{zhen2020smap}, we show the qualitative results of our method compared with that of the two SOTA methods RootNet (top-down) and SMAP (bottom-up) in Fig \ref{fig:mupots}. 


\paragraph{Additional Comparison on Wild Videos}

\begin{figure*}
    \centering
    \includegraphics[width=15cm]{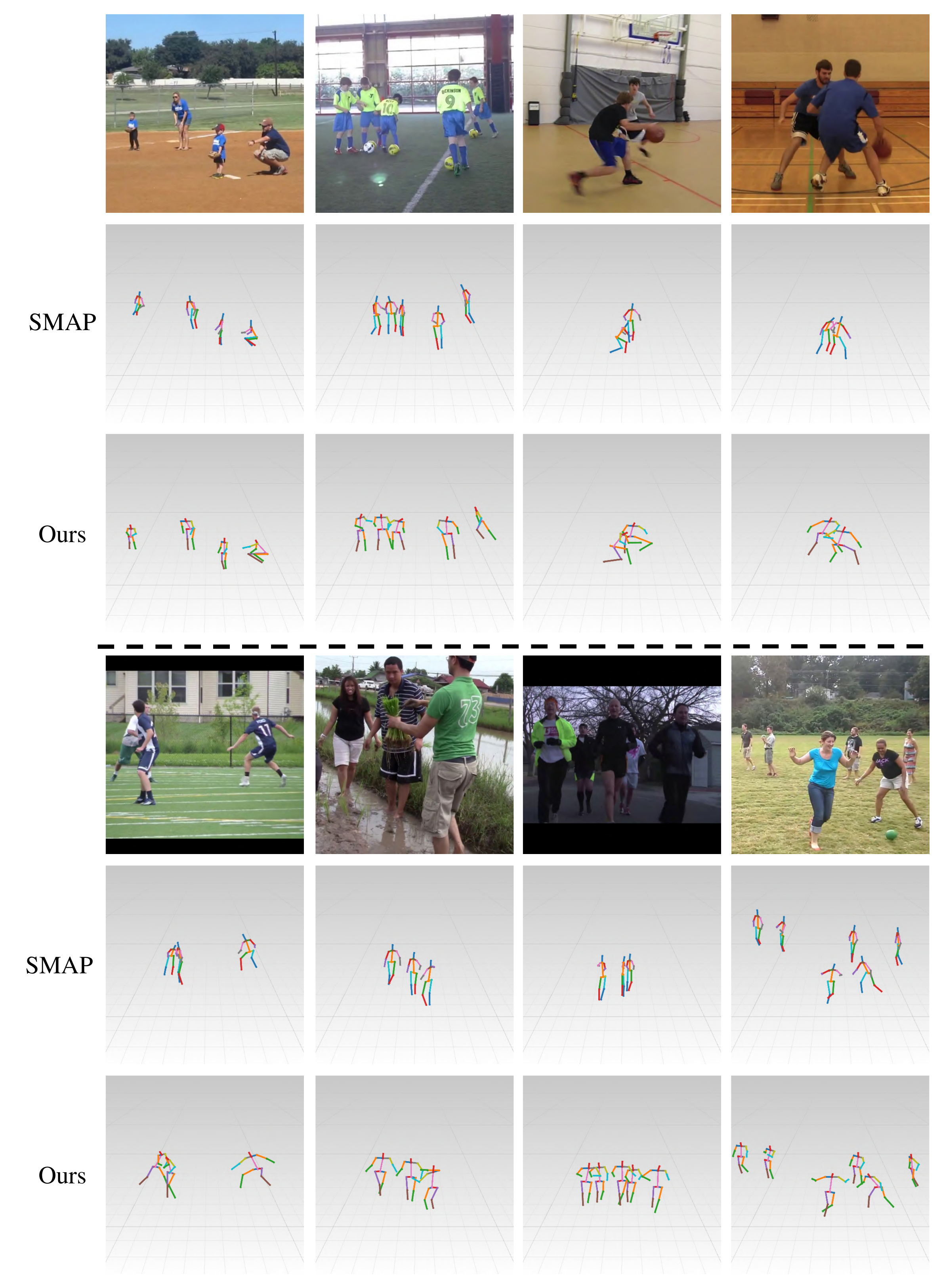}
    \caption{Results of our method compared with that of SMAP \cite{zhen2020smap} (i.e., the SOTA bottom-up method) on wild videos. Results from eight video clips are included (i.e., one frame for each video). Four results are at top part of the figure, the other four are at the bottom, separated by the dashed line. For either part, the first row is the frames from the video clip; the second row is the results of SMAP; the third row is the results of our method. These results again show that the SOTA method cannot handle inter-person occlusions. In contrast, our method produces accurate camera-centric 3D multi-person pose estimation. }
    \label{fig:wild}
\end{figure*}

To further demonstrate the performance of our method compared with the SOTA 3D multi-person pose estimation method. We provide the qualitative results of our method compared with that of the SOTA bottom-up method SMAP \cite{zhen2020smap} in Fig \ref{fig:wild}. The video clips are selected from MPII \cite{andriluka14cvpr} dataset which is neither used for training or evaluation for both methods. 

\paragraph{Additional Comparison on JTA}


\begin{figure*}
    \centering
    \includegraphics[width=0.9\textwidth]{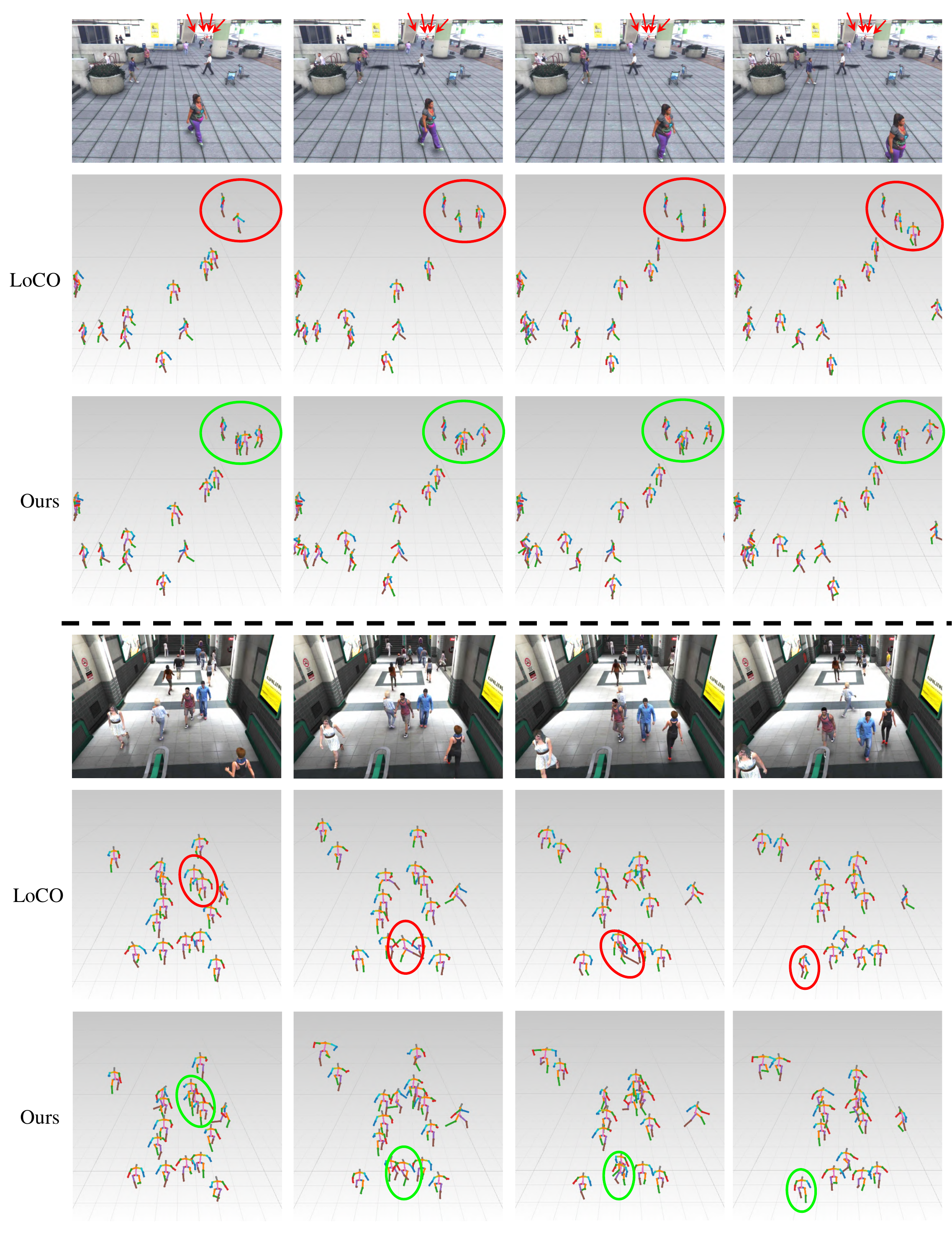}
    \caption{Result of our method compared with that of LoCO \cite{fabbri2020compressed} (i.e., a SOTA method released trained model on JTA) on JTA dataset. Results from two video clips are included: top and bottom separated by the dashed line. For each video clip, the first row is the frames from the video clip; the second row is the result of LoCO; the third row is the result of our method. These results show that on this synthetic datasets, our method is able to produce more accurate and robust 3D multi-person pose estimation compared with other methods. We use \textcolor{red}{red} circle to indicate the wrong results of LoCO and \textcolor{green}{green} circle to point out the corresponding correct results of our method. In the first row of the top video clip, due the four persons are far from the camera which are small, we use four \textcolor{red}{red} arrows to indicate each of them.}
    \label{fig:jta}
\end{figure*}

As we reported our quantitative performance on JTA dataset in Table 4 in the main paper, we also provide the qualitative results of our method compared with that of the SOTA method reported and released their trained model on the JTA dataset \cite{fabbri2020compressed} in Fig \ref{fig:jta}. The two video clips in Fig \ref{fig:jta} show both inter-person occlusions and large multi-person scale variation where we observe our method can handle both challenges well and produce accurate camera-centric 3D multi-person pose estimation compared with LoCO \cite{fabbri2020compressed}.